\documentclass[10pt, journal]{IEEEtran}

\usepackage{amsmath,epsfig}
\usepackage{amsfonts}
\usepackage{times}
\usepackage{balance}
\usepackage{amssymb}
\usepackage{graphicx}
\usepackage{euscript}
\usepackage{algorithm, algorithmic}
\usepackage{color}
\usepackage{breqn}
\usepackage{subfigure}
\usepackage{multirow,hhline}
\usepackage{tabularx,colortbl}
\usepackage{placeins}
\usepackage{comment}

\newcommand{\ru} {\rule{0mm}{3mm}}

\definecolor{blue(ryb)}{rgb}{0.00, 0.50, 1.00}
\renewcommand{\r}[1]{{\color{red} {#1}}}
\renewcommand{\b}[1]{{\bf \color{blue(ryb)}{#1}}}

\begin{document}
\title{A Full-Image Full-Resolution End-to-End-Trainable CNN Framework for Image Forgery Detection}

\author{Francesco~Marra, Diego~Gragnaniello, Luisa Verdoliva and Giovanni~Poggi
\IEEEcompsocitemizethanks{
\IEEEcompsocthanksitem F.Marra, D.Gragnaniello and G.Poggi are with the DIETI, L.Verdoliva is with the DII,
Universit\`{a} degli Studi di Napoli Federico II, Naples, Italy.
E-mail: \{francesco.marra, diego.gragnaniello, verdoliv, poggi\}@unina.it.}}

\IEEEcompsoctitleabstractindextext{%
\begin{abstract}
Due to limited computational and memory resources, current deep learning models accept only rather small images in input, calling for preliminary image resizing.
This is not a problem for high-level vision problems, where discriminative features are barely affected by resizing.
On the contrary, in image forensics, resizing tends to destroy precious high-frequency details, impacting heavily on performance.
One can avoid resizing by means of patch-wise processing, at the cost of renouncing whole-image analysis.

In this work, we propose a CNN-based image forgery detection framework which makes decisions based on full-resolution information gathered from the whole image.
Thanks to gradient checkpointing, the framework is trainable end-to-end with limited memory resources and weak (image-level) supervision, allowing for the joint optimization of all parameters.
Experiments on widespread image forensics datasets prove the good performance of the proposed approach, which largely outperforms all baselines and all reference methods.
\end{abstract}
\begin{IEEEkeywords}
Digital image forensics, CNN, forgery detection.
\end{IEEEkeywords}
}
\maketitle

\IEEEdisplaynotcompsoctitleabstractindextext
\IEEEpeerreviewmaketitle

\section{Introduction}

In this work, we propose a new framework for image forgery detection based on convolutional neural networks (CNN).
This may not look particularly exciting: deep learning is by-now common practice to solve all kinds of vision-related problems.
However, image forensics has some peculiarities that set it apart from standard computer vision problems.
We can summarize them in the need to look, at the same time, at the whole image but also at its tiniest details.
Consider the example of Fig.1.
This well-crafted splicing does not show obvious artifacts that allow detection by visual inspection,
but a suitable structural analysis reveals differences that may be due only to the insertion of alien material in the host image.
Indeed, many state-of-the-art forensic tools rely on the statistical analysis of local micro-patterns.
However, {\em local} analyses alone are necessarily suboptimal.
Clues emerging from the whole image, and at multiple scales, should be combined and processed jointly to make a reliable decision.
Therefore, our goal is to design CNN-based forensic tools that, overcoming current technological limitations,
meet the contrasting requirements of full-resolution and full-image training and analysis.

It should be realized that this problem is indeed peculiar of multimedia forensics.
Typical CNN classifiers for computer vision problems rely on {\em macroscopic} features, which bear high-level semantic clues on the scene.
For example, a face detector may look for the presence of specific facial features with suitable spatial relationships.
Such large-scale information persists nicely after resizing the image.
And in fact, target images of wildly different sizes are routinely resized to match the input CNN layer.
Actually, resizing is even used on purpose, during training, to gain robustness to scale changes.
In the context of image forensics, instead, resizing may destroy the very same information classifiers rely upon,
the pixel-level {\em micro-patterns} that characterize different digital histories.
By analyzing such patterns one can identify camera models, individual devices, or discover the traces of out-camera processing.
A huge scientific literature testifies on the importance of such high-frequency features.
Hence, image resizing and resampling should be definitely avoided when performing forensic tasks.

\begin{figure}
	\centering
	\setlength{\tabcolsep}{1mm}
	\begin{tabular}{c}
	\includegraphics[width=0.99\linewidth]{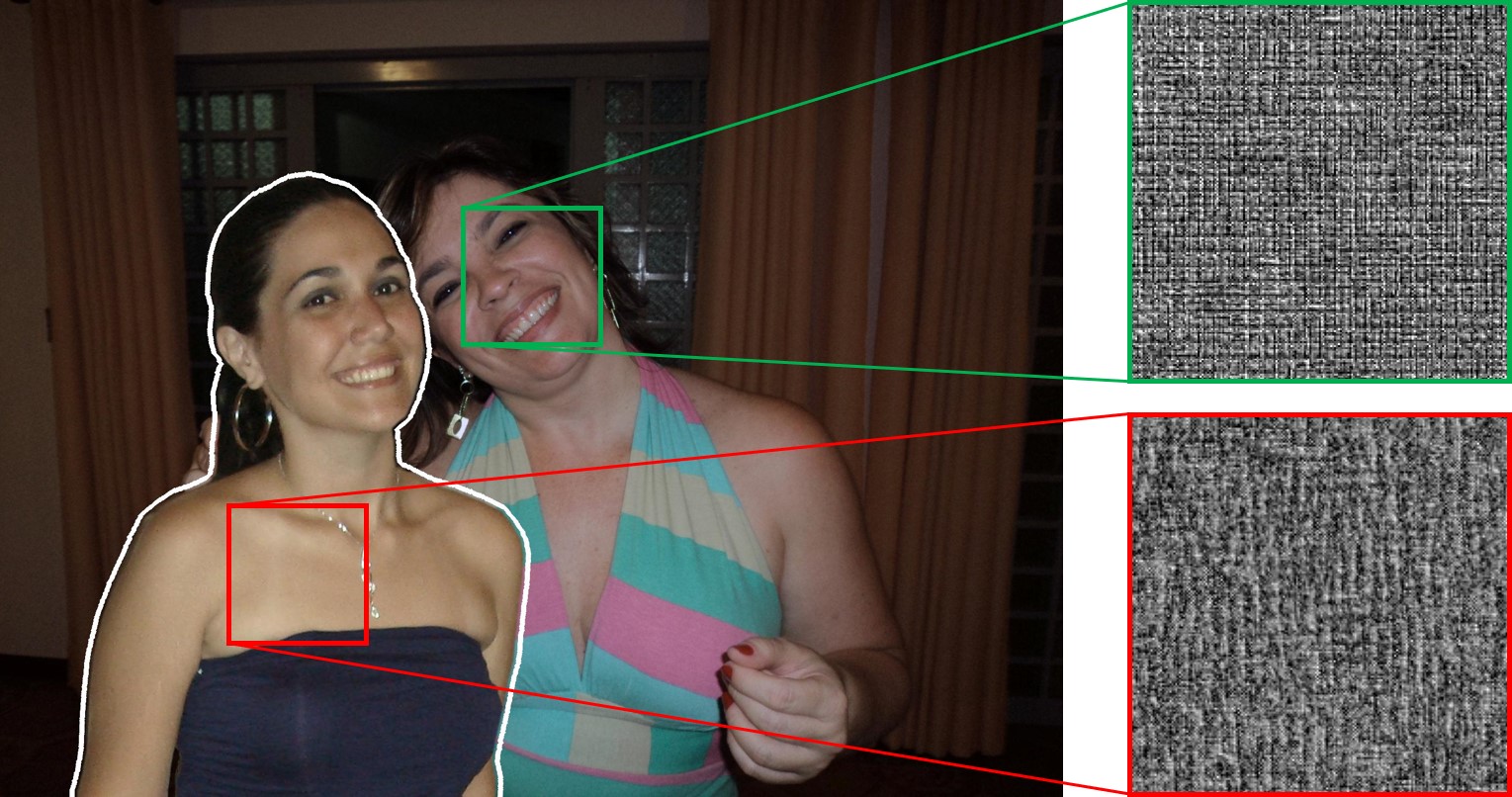}
    \end{tabular}
	\caption{Example of carefully crafted splicing. Visual inspection does not allow detection, but pixel-level analyses expose suspicious textural differences.}
	\label{fig:teaser}
\end{figure}

So, one could naively think of using a network with an input size as large as the target image.
Besides the lack of generality (images can be of any size) a more fundamental issue concerns computational and memory resources.
Acquisition devices are continuously improving their resolution, with commercial smart-phone cameras delivering photos with many millions of pixels.
Deep learning hardware capabilities do not increase at the same rate.
Due to computation and memory limitations, state-of-the-art architectures accept only small images in input, especially when very deep networks are used.
Therefore, the highly informative image samples cannot be directly fed to a network and analyzed as a whole.

Eventually, when high-resolution must be preserved,
a simple solution is to perform patch-wise feature extraction, followed by some forms of feature aggregation to exploit the full-image information.
This approach makes full sense, and largely predates deep learning.
Yet, even with good CNN-based feature extractors and classifiers, it is inherently suboptimal for several reasons:
{\it   i)} poor feature extraction;
{\it  ii)} poor global decision;
{\it iii)} need of over-detailed ground truth.

First of all, since the patch-wise feature extractor is trained without taking into account whole-image information,
the best it can do is to learn good features for {\em local} decisions, which are not necessarily the best ones in view of future aggregation.
Then, the global classifier, trained after freezing the patch-level processing, operates only on intermediate features,
hence is necessarily suboptimal with respect to a classifier trained end-to-end on the original data.
Last, patch-wise training requires a detailed, handcrafted, ground truth.
Therefore,
the large datasets necessary to train deep learning models require a huge man-power and are inevitably affected by errors, with a sure impact on the eventual performance.

All these considerations motivate our work, and allow us to define the final goal more clearly.
We want to design deep learning models for image forgery detection which are:
\begin{enumerate}
\item   full-image: make decisions based on information gathered from all over the image;
\item   full-resolution: do not perform any harmful image resizing;
\item   end-to-end trainable: optimize jointly all model parameters for image-level classification, based only on image-level (weak) supervision.
\end{enumerate}
To achieve this goal, we propose a framework comprising three blocks in cascade performing, respectively,
patch-wise feature extraction, image-wise feature aggregation, and global decision.
All blocks are fully trainable, based on image-wise labels, allowing information to flow backward through the whole network.
The global decision takes into account features extracted from the whole image, whatever its size, and based on local micro-patterns.
Memory problems are solved by means of gradient checkpointing, with a very limited increase of computational costs.

With these solutions, the proposes framework allows one to optimize jointly
the local information extraction, the global feature aggregation, and the whole-image classification, whatever the input image size.
We implemented several versions of this general framework, through appropriate selection of the major architectural blocks.
After training on suitable synthetic datasets, we performed extensive experiments on realistic datasets widespread in the image forensics community,
focusing on local manipulations, such as splicings, copy-moves, and inpainting, likely indicators of malicious attacks.
Results fully support our approach
which largely outperforms both baseline methods and state-of-the-art references, including methods requiring strong supervision.

In the following, we
analyze related work (Section II),
describe the proposed approach (Section III),
report on the results of numerical experiments (Section IV),
and finally draw conclusions (Section V).

\section{Related work}

Forgery detection is a central topic in image forensics, and there is a large bulk of relevant literature.
In addition, it is necessary to consider both forgery detection and localization, since these tasks are tightly related.
Indeed, detection methods can be used for localization through sliding-window analysis, and localization method may allow detection by suitable post-processing.
So, to limit the scope, in the following analysis we take a historical perspective, but focus especially on recent CNN-based methods.
Moreover, we neglect global manipulations, such as histogram equalization or gamma correction, as they can be hardly regarded as malicious forgeries.

Early contributions were mostly model-based, looking for statistical anomalies related to
the color filter array (CFA) \cite{Mahdian2009, Ferrara2012}, double JPEG compression \cite{Ye2007, Bianchi2012}, or sensor noise \cite{Chen2008, Chierchia2014}.
Most of these methods assume {\it a priori} the presence of a forgery
and pursue localization through pixel-level analysis, generating a heat-map.
Then, a global score can be easily computed from the latter and used for detection.
Model-based approaches are elegant and do not require extensive training, but work only in quite restrictive hypotheses.

The advent of data-driven solutions granted a quantum leap in performance and ensured higher generality.
Methods based on machine learning extract suitable hand-crafted features from the image,
both in the spatial domain \cite{Kirchner2010,Cozzolino2014a,Zhao2015,Cozzolino2015,Li2018} and in the transform (DCT, wavelet) domain \cite{Lyu2005,Shi2008,He2012},
which are used to train a classifier.
Extracting features from the whole image allows direct and reliable image forgery detection.
Instead, localization can be obtained by working in sliding-window modality and using a suitable local score.
The most discriminative features rely on high-order image statistics which help revealing spatial inconsistencies originated by the presence of forgeries.
To this end, high-pass residual images are often used, obtained by means of derivative filters \cite{Fridrich2012} or image denoisers.

In recent years, methods based on deep learning have become dominant.
Some early papers, inspired by the success of residual-based machine learning methods,
propose CNN architectures with a first layer of high-pass filters, either fixed \cite{Rao2016,Liu2018}, or trainable \cite{Bayar2016}, meant to extract residual feature maps.
In \cite{Cozzolino2017} it is even shown that successful methods based on hand-crafted features can be recast as CNNs and fine tuned for improved performance.
In \cite{Zhou2018} these low-level features are augmented with high-level ones in a two-stream CNN architecture.
Recent findings \cite{Marra2018, Roessler2019}, however,
show that such constrained first layer is only useful with small networks and datasets.
Given a suitably large training set, general-purpose very deep architectures provide the same good results in favourable cases,
but ensure higher robustness to compression and training/test misalignments.

Several papers, to begin with \cite{Rao2016}, followed more recently by \cite{Salloum2018} and \cite{Zhang2018},
train explicitly the net to distinguish between homogeneous and heterogeneous patches, the latter characterized by the presence of both pristine and forged areas.
The rationale is to catch the patterns that characterize transitions regions, anomalous with respect to the background,
so as to localize possible forgeries.
This idea is followed also in \cite{Bappy2019}, where an hybrid CNN-LSTM architecture is trained end-to-end to produce a binary mask for forgery localization.
These methods, however, require detailed ground truth maps to train the net, which may not be available or precise.

For architectural constraints, most of these methods carry out a patch-based analysis, working on relatively small patches,
with further steps needed to compute a global score at image-level.
In \cite{Rao2016}, for example, the CNN extract features patch-wise and later aggregates them in a global feature vector used to feed a SVM classifier.
This may impact on detection performance.
A more fundamental limit concerns the need of strongly aligned training and test sets.
Some methods, {\it e.g.}, \cite{Zhang2018,Bappy2019}, carry out experiments on a single database split into training and test,
others \cite{Zhou2018} require fine-tuning on target data.
All this highlights the limited generalization ability of supervised learning, as also shown in \cite{Cozzolino2018FT}.

A more promising line of research is to revisit the anomaly detection approach under a data-driven paradigm.
Anomalies are detected by means of single-image analyses, with a sort of blind source identification.
In \cite{Cozzolino2016} this was accomplished in a fully unsupervised fashion by using an autoencoder architecture.
More recent proposals \cite{Bondi2017,Cozzolino2019,Cozzolino2018} use camera-model features, gathered off-line by dedicated CNNs,
or leverage metadata information \cite{Huh2018} for direct detection.
A strong pro of this approach is that training is performed only on pristine images, with no need of aligned datasets and ground truths,
which ensures good robustness and adaptability to unseen manipulations.
In \cite{Cozzolino2019} and \cite{Huh2018}, in particular,
this is achieved by using a Siamese training on pairs of patches extracted from pristine images, with a suitable consistency metric.

Besides its technical content,
this short review of ideas makes clear that there is high and growing interest for new solutions in this field,
to face the threats posed by increasingly sophisticated fake multimedia tools.

\section{Proposed method}

\begin{figure}[t!]
	\centering
	\setlength{\tabcolsep}{1mm}
	\begin{tabular}{c}
		\includegraphics[width=0.99\linewidth]{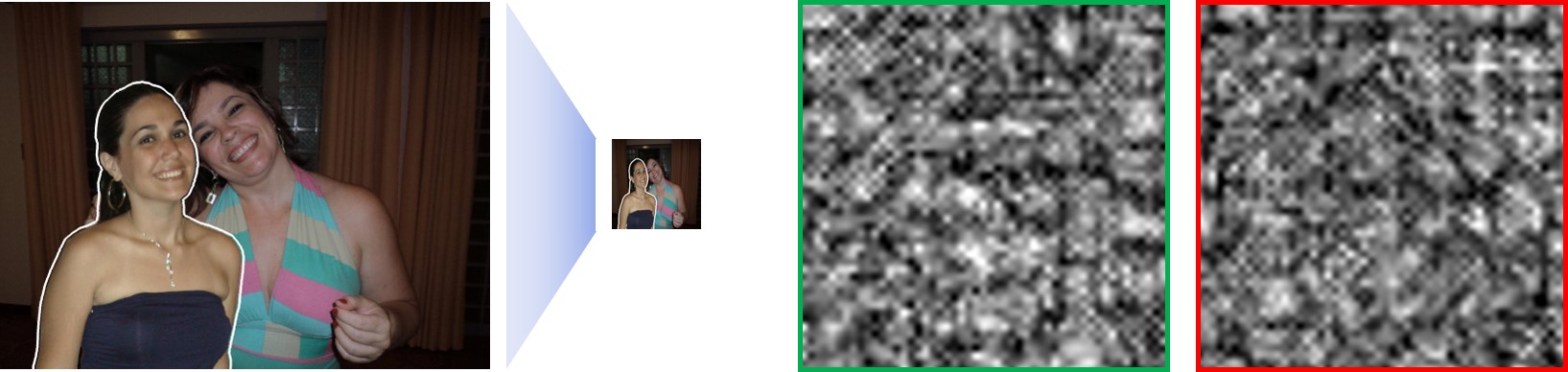}
	\end{tabular}
	\caption{Strong image resizing corrupts the textural patterns used in forensics.}
	\label{fig:teaser_resized}
\end{figure}

Our aim is to design a deep network to detect the presence of localized forgeries in a target image, irrespective of the image size and the forgery size.
Of course, images can have wildly different sizes, depending also on the context, but the trend is towards higher and higher resolutions.
Today's smartphones feature cameras with resolutions of 10 Mpixels and more.
On the other hand, due to computation/memory bottlenecks, deep networks accept rather small images in input, for example 256$\times$256-pixel.
Hence, a strong size mismatch typically occurs between target image and network input.
For most image analysis applications, this mismatch is not a big problem and two solutions can be considered:
\begin{enumerate}
\item   images are rescaled to fit the network input, or
\item   images are processed patch-wise, and results are fused off-line to make a global decision.
\end{enumerate}
In the following paragraphs,
we first explain why such solutions are not viable for image forgery detection,
then describe the proposed architecture,
and finally show how it can be trained end-to-end based on the gradient checkpointing method.

\begin{figure*}
    \centering
    \includegraphics[width=0.9\textwidth]{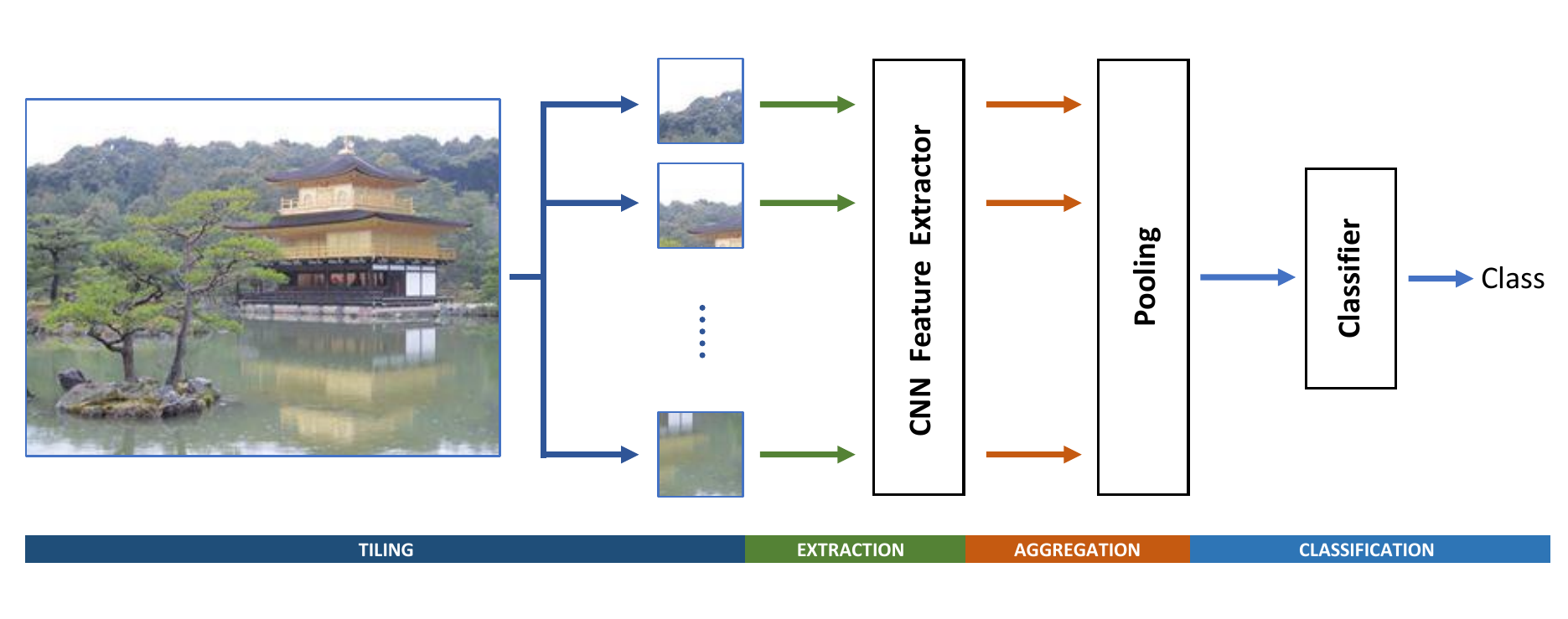}
    \caption{Proposed end-to-end trainable framework for image forgery detection, comprising extraction, aggregation, and classification blocks.}
    \label{fig:framework}
\end{figure*}

\subsection{The need for full-image full-resolution processing}

The first solution listed before is to rescale the image to fit the network first layer.
However, this is not advisable when dealing with forgery detection.
In some cases, the forged region could be so small to become practically undetectable after strong downsampling.
A more fundamental problem, however, is that
some sophisticated forgeries may only be detected based on the statistical analyses of micro-textures.
However, these precious high-frequency components are strongly corrupted when the image is resized or resampled.
Fig.\ref{fig:teaser_resized} shows a clear example,
in which the markedly different textures highlighted in Fig.\ref{fig:teaser},
after resizing become very similar to one another and basically useless for forensic analyses.

The second solution is to perform patch-level detection, with no resampling, followed by some form of information fusion to make a global decision.
Indeed, given an ideal patch-level classifier, the fusion problem has an obvious solution,
and the presence of a forgery can be declared if at least one forged patch is detected.
However, real-world detectors are far from ideal, they always have non-zero missing-detection and false-alarm rates.
For example, assuming a rather optimistic 1\% patch-level false-alarm rate, and independent decisions,
a 100-patch pristine image would present a false-alarm rate beyond 63\%.
Therefore, the fusion problem is not at all trivial with real-world detectors, as our experiments will confirm.
In addition, the patch-level detector itself should be designed taking into account image-level performance.

These considerations motivate the need for a full-image full-resolution detector.
In this way, precious microtextures can be preserved and, at the same time, information coming from all patches can be processed jointly to make a reliable decision.
A naive implementation of this idea, with a CNN input size matching the image size, would require huge computational and memory resources,
not to speak of the number of images needed for reliable training.
Instead, we propose a suitable architecture that, through reasonable structural constraints, satisfies the needs of forensics detection with limited resources.

\subsection{Proposed architecture}

The proposed framework is represented pictorially in Fig.\ref{fig:framework}.
It consists of three blocks performing, respectively, patch-level feature extraction, feature aggregation, and decision.
Note that, although we propose a specific implementation of such blocks, this is not the core of our proposal, which is instead the whole framework.

\subsubsection{Patch-level feature extraction}
after dividing the image in overlapping patches, these are processed to extract discriminative features.
As feature extractors, we adopt some state-of-the-art deep networks,
taking the output of the penultimate layer as feature vector, and discarding the final class probabilities.
However, considering the peculiarities of image forgery detection,
we modify the input layer to accommodate some additional inputs, the image noiseprint \cite{Cozzolino2018}, besides the image color bands.
Noiseprints are high-pass image residuals, extracted through a dedicated network, in which camera-related artifacts are emphasized.
Therefore, they highlight possible spatial anomalies and may help detecting local manipulations.

\subsubsection{Feature aggregation}
the feature extractor produces a large number of features,
which are aggregated image-wise to obtain a single descriptor for the classification task.
To this end, we consider several forms of pooling, maximum, minimum, average, and average of squares:
\begin{equation}
\begin{split}
    F_{\rm max}  & = \max_{i=1,..,N_p} F_i \\
    F_{\rm min}  & = \min_{i=1,..,N_p} F_i \\
    F_{\rm mean} & = \frac{1}{N_p} \textstyle{\sum_{i=1}^{N_p}} \, F_i \\
    F_{\rm msq}  & = \frac{1}{N_p} \textstyle{\sum_{i=1}^{N_p}} \, F_i^2
\end{split}
\label{eq:aggregations}
\end{equation}
where $F_i=[F_{i,1},\ldots,F_{i,C}]$ is the $C$-component feature extracted from the $i$-th patch, $N_p$ is the number of (possibly overlapping) patches, and all operations on features are component-wise.
The most appropriate type of pooling depends on the problem of interest.
When the information is spread over the whole image, an average pooling is reasonable,
while min or max pooling are more appropriate when the discriminative information is concentrated in a localized region.
In any case, we also use the combination of multiple types of pooling, leaving the final choice to experiments.
After aggregation all explicit spatial dependencies are discarded.

\newcommand{\pdev}[2]{{\frac{\partial #1}{\partial #2}}}
\renewcommand{\L}{{\cal L}}
\newcommand{\rup}{\rule{0mm}{10mm}}
\newcommand{\agg}{{\rm agg}}
Note that the type of pooling impacts on how information back-propagates from the output to update the parameters of the feature extractor.
In more detail,
let $F_\agg$ denote the aggregated feature, $\L$ the loss function of the framework, and $\theta$ a generic parameter of the CNN.
Then, the gradient of $\L$ with respect to $\theta$ reads
\begin{equation}
    \pdev{\L}{\theta} = \sum_{c=1}^C \pdev{\L}{F_{\agg,c}} \pdev{F_{\agg,c}}{\theta}
\end{equation}
with
\begin{equation}
    \pdev{F_{\agg,c}}{\theta} =
        \begin{cases}
        \rup \displaystyle   \pdev{F_{i,c}}{\theta} \cdot \delta_{i,i_{\max}(c)}            & \text{max pooling} \\
        \rup \displaystyle   \pdev{F_{i,c}}{\theta} \cdot \delta_{i,i_{\min}(c)}            & \text{min pooling} \\
        \rup \displaystyle   \frac{1}{N_p}\sum_{i=1}^{N_p}         \pdev{F_{i,c}}{\theta}   & \text{average pooling} \\
        \rup \displaystyle   \frac{1}{N_p}\sum_{i=1}^{N_p} 2F_{i,c}\pdev{F_{i,c}}{\theta}   & \text{av.square pooling} \\
\end{cases}
\label{eq:backprop_agg}
\end{equation}
In the above equation, $\delta_{i,j}$ equals 1 when $i$=$j$ and 0 otherwise, while $i_{\max}(c)$ and $i_{\min}(c)$ point to the feature vectors with the largest, respectively smallest, $c$-th component.
Therefore,
with max or min pooling, only some ``active'' patches contribute to the gradient, and are updated during training.
Instead, with mean and msq pooling all patches are involved.
Of course, when multiple forms of pooling are used at the same time, the gradient is obtained as the weighted sum of the individual terms.

\subsubsection{Decision}
after aggregating the local information in a single descriptor $F$ for the whole image, this is classified by means of a few fully-connected layers.
This is the typical classifier used in deep networks, and usually two layers provide a good trade-off between complexity and accuracy.

\subsection{End-to-end training}

If we focus only on the post-training operations,
the proposed architecture does not look much different from conventional approaches based on patch-wise feature extraction, pooling, and classification.
Contrary to such approaches, however, our framework is trainable {\em end-to-end}.
This means that we do not train the feature extractor on individual labeled patches and, afterwards, train the classifier on the features extracted by a fixed net.
Instead, we train the whole framework, top to bottom, on full-size images, with a single label associated with each one: forged or pristine.
The loss back-propagates through the net up to to individual patches,
allowing the feature extractor to learn which information is the most discriminative for the final decision,
and adapting the classifier jointly with the extractor itself.

To better underline the difference with respect to patch-wise training,
consider that in a large image with a localized forgery most patches are actually pristine, and only a few ones truly forged.
In our end-to-end training, all these patches share the same image-level label (forged).
Therefore, the net is forced to learn how to manage such contrasting indications to make the correct decision.
As a side benefit, there is no need to have a pixel-wise ground truth for training, since the only relevant label applies to the whole image.
Also, images of any size can be used for training, with forgeries of any size (especially if max/min pooling is used).

Going into technical details
for each training batch of images, the framework performs
{\it  i)} an inner loop on the patches of each image, computing the back-propagation at the end of the loop, and
{\it ii)} an outer loop on the images of the batch, that sums up gradients computed for each inner loop and finally updates the weights once at the end of the batch.
Due to the arbitrary size of input images, each inner loop involves a different number of patches,
impacting on the computational effort, which may vary significantly from batch to batch.
This is a minor issue, though, with respect to memory requirements.
In fact, to back-propagate the loss, gradients must be computed for all processed patches, causing an increase of the occupied memory, which grows linearly with the image size.
For deep networks and large images, this memory is simply unavailable.

The situation is described pictorially in Fig.4,
where a circle represents a layer, and a black dot at the center indicates that activations are stored.
In the forward pass (a), in fact, all activations at each layer are computed and stored.
Then, in the backward pass (b), they are used to propagate gradients from the last layer, where the loss is computed, to the input.
After usage, they are erased (small dots).
It should be realized that deep nets can include hundreds of layers, with several feature maps at each layer, whose size is typically proportional to the input size.
Therefore, to process a large input image at once, a huge number of variables should be stored, exceeding the available memory.

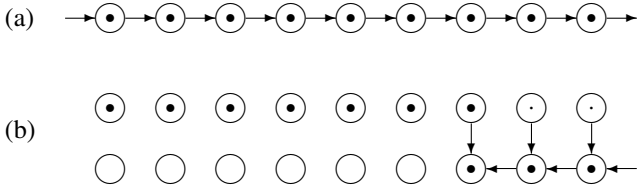
\begin{figure}
\setlength{\unitlength}{1mm}
\noindent
\begin{picture}(88,25)(00,00)

\put(00,21){(a)}
\multiput(08,22)(8,0){10}{\vector(1,0){4}}
\multiput(14,22)(8,0){09}{\circle{4}}
\multiput(14,22)(8,0){09}{\circle*{1}}

\put(00,06){(b)}
\multiput(14,10)(8,0){09}{\circle{4}}
\multiput(14,10)(8,0){07}{\circle*{1}}
\multiput(70,10)(8,0){02}{\circle*{0.2}}
\multiput(62,08)(8,0){03}{\vector(0,-1){4}}
\multiput(68,02)(8,0){03}{\vector(-1,0){4}}
\multiput(14,02)(8,0){09}{\circle{4}}
\multiput(62,02)(8,0){03}{\circle*{1}}

\end{picture}
\label{fig:chechpointing}
\caption{Conventional CNN training with backpropagation.}
\end{figure}

To manage this problem we resort to the gradient checkpointing strategy, originally proposed in \cite{Chen2016}, which trades off memory for computation.
This solution is described pictorially in Fig.5.
During the forward pass (a), all activations are deleted immediately after use, except for those in a few ``checkpoint'' layers (red dots).
In the backward pass (b)-(e), gradients are computed one group at a time (in the figure we show two groups of 4 layers).
Since activations are necessary to this end, they are recomputed, but only from the last checkpoint on, (b).
This allows backpropagating the gradient until the checkpoint layer itself (c).
At this point all variables at layers beyond the checkpoint are deleted, and the process goes on with a new group of layers (d)-(e).

With a judicious choice of the number of checkpoints, memory occupation can be significantly reduced and become manageable.
Of course, each activation is computed twice, but the computational overhead is limited, because the forward pass is lighter than the backward pass.
Note that gradient checkpointing has been recently made available in PyTorch as well as in other platforms.
With this solution, we were able to train our network end-to-end seamlessly, with a increase of the training time that never exceeded 20\%.

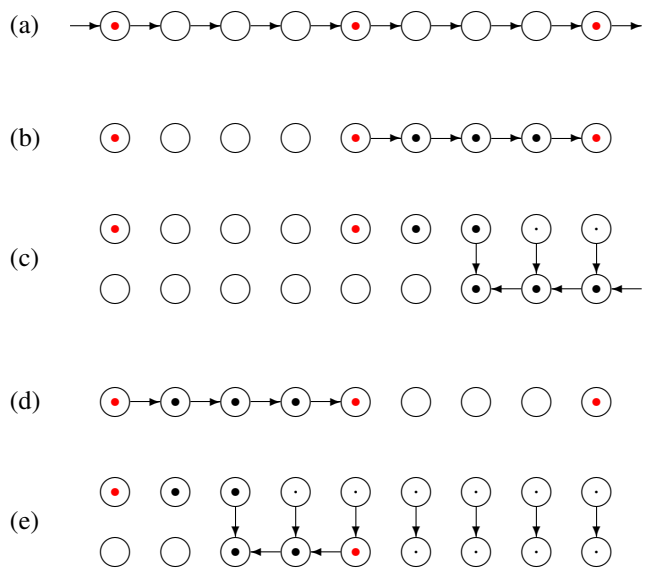
\begin{figure}
\setlength{\unitlength}{1mm}
\noindent
\begin{picture}(88,78)(00,00)

\put(00,073){(a)}
\multiput(08,74)(8,0){10}{\vector(1,0){4}}
\multiput(14,74)(8,0){09}{\circle{4}} {\color{red}
\multiput(14,74)(32,0){3}{\circle*{1}} }

\put(00,058){(b)}
\multiput(48,59)(8,0){04}{\vector(1,0){4}}
\multiput(14,59)(8,0){09}{\circle{4}} {\color{red}
\multiput(14,59)(32,0){3}{\circle*{1}} }
\multiput(54,59)(8,0){03}{\circle*{1}}

\put(00,042){(c)}
\multiput(14,47)(8,0){09}{\circle{4}} {\color{red}
\multiput(14,47)(32,0){2}{\circle*{1}} }
\multiput(54,47)(8,0){02}{\circle*{1}}
\multiput(70,47)(8,0){02}{\circle*{0.2}}
\multiput(62,45)(8,0){03}{\vector(0,-1){4}}
\multiput(68,39)(8,0){03}{\vector(-1,0){4}}
\multiput(14,39)(8,0){09}{\circle{4}}
\multiput(62,39)(8,0){03}{\circle*{1}}

\put(00,023){(d)}
\multiput(16,24)(8,0){04}{\vector(1,0){4}}
\multiput(14,24)(8,0){09}{\circle{4}} {\color{red}
\multiput(14,24)(32,0){3}{\circle*{1}} }
\multiput(22,24)(8,0){03}{\circle*{1}}

\put(00,007){(e)}
\multiput(14,12)(8,0){09}{\circle{4}} {\color{red}
\multiput(14,12)(32,0){1}{\circle*{1}} }
\multiput(22,12)(8,0){02}{\circle*{1}}
\multiput(38,12)(8,0){06}{\circle*{0.2}}
\multiput(30,10)(8,0){07}{\vector(0,-1){4}}
\multiput(36,04)(8,0){02}{\vector(-1,0){4}}
\multiput(14,04)(8,0){09}{\circle{4}}
\multiput(30,04)(8,0){02}{\circle*{1}}{\color{red}
\multiput(46,04)(8,0){01}{\circle*{1}} }
\multiput(54,04)(8,0){04}{\circle*{0.2}}

\end{picture}
\label{fig:chechpointing}
\caption{CNN training with gradient checkpoints.
After the forward pass (a), activations are stored only at checkpoint layers (red). The backward pass (b)-(e) proceeds one group of layers at a time. Activations at intermediate layers must be recomputed each time a group is processed.}
\end{figure}

\section{Experimental analysis}

In this Section we design and perform numerical experiments to assess the performance of the proposed approach.

In the following subsections, we
first describe the training procedure,
then present the results of some preliminary experiments carried out to make key design choices,
and finally compare the proposed method with both baselines and state-of-the-art references
on several challenging datasets widespread in the community.

\subsection{Training}

\setlength{\tabcolsep}{6pt}
\begin{table*}
\centering
\caption{Features of datasets used for training and testing}
\begin{tabular}{|l|l|l|c|c|c|} \hline
   \ru dataset           & manipulations        & counter forensic                         & \# prist. / forged &                          resolution   & format    \\ \hline
   \ru Vision / UCID     & automatic splicing   & -                                        &    7565 / $\infty$ &  960$\times$720  $-$ 4640$\times$3480 &  JPG      \\ \hline
   \ru Dresden / FAU     & automatic splicing   & -                                        &       4992 / 14976 & 2560$\times$1920 $-$ 4352$\times$3264 &  JPG      \\ \hline
   \ru DSO-1             & splicing             & color/contrast adjustment                &        100 / 100   & 2048$\times$1536                      &  PNG      \\ \hline
   \ru Korus             & splicing, copy-move  & -                                        &        220 / 220   & 1920$\times$1080                      &  TIF      \\ \hline
   \ru                   & splicing, copy-move, & color/contrast adjustment, PRNU editing, &                    &                                       &  PNG, BMP \\
   \ru NC2017            & computer-generated,  & JPEG quantization, cloning               &       2470 / 1051  &  436$\times$600  $-$ 3648$\times$5472 &  JPG      \\
   \ru                   & inpainting           &                                          &                    &                                       &           \\ \hline
   \ru                   & splicing, copy-move, & color/contrast adjustment, PRNU editing, &                    &                                       &  PNG, BMP \\
   \ru MFC2018           & computer-generated,  & JPEG quantization, cloning, noising,     &      12246 / 1935  &  352$\times$512  $-$ 5470$\times$7586 &  JPG, TIF \\
   \ru                   & inpainting           & dithering, social network laundering     &                    &                                       &           \\ \hline
\end{tabular}
\label{tab:datasets}
\end{table*}

In order to train our networks, we generated a suitable synthetic dataset.
Background images are taken from the Vision dataset, proposed originally \cite{Shullani2017} for camera model identification,
which comprises $7565$ images acquired by $35$ different devices with the native high-quality JPEG compression.
To generate manipulated images, we spliced on them objects drawn
from a set of 81 objects manually cropped from the uncompressed images of the UCID dataset \cite{Schaefer2003}.
Details on all datasets used in this work are reported in Tab.\ref{tab:datasets}.

\begin{figure}[t]
	\centering
	\setlength{\tabcolsep}{1mm}
	\begin{tabular}{cc}
	\includegraphics[width=0.45\linewidth]{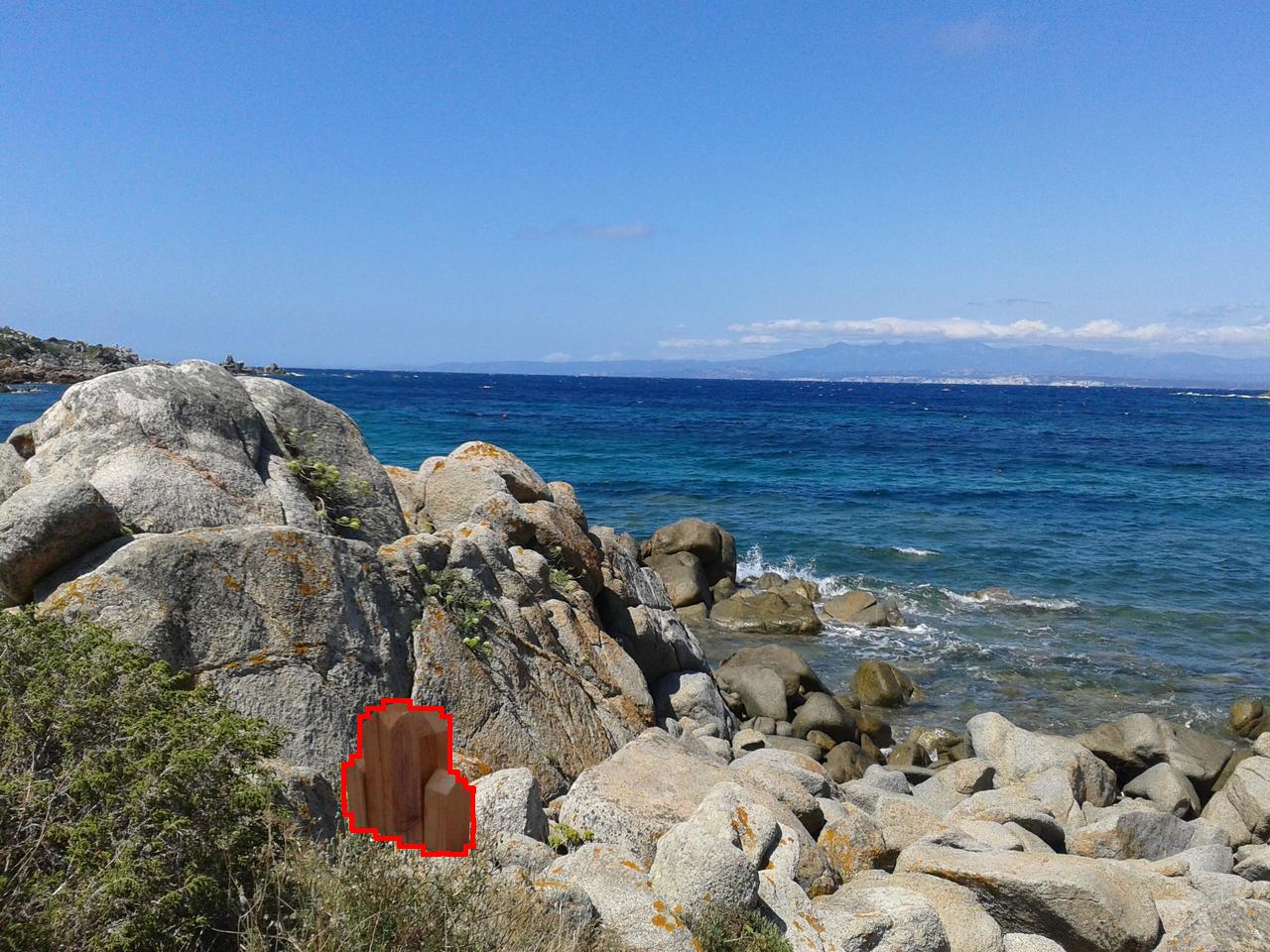} &
	\includegraphics[width=0.45\linewidth]{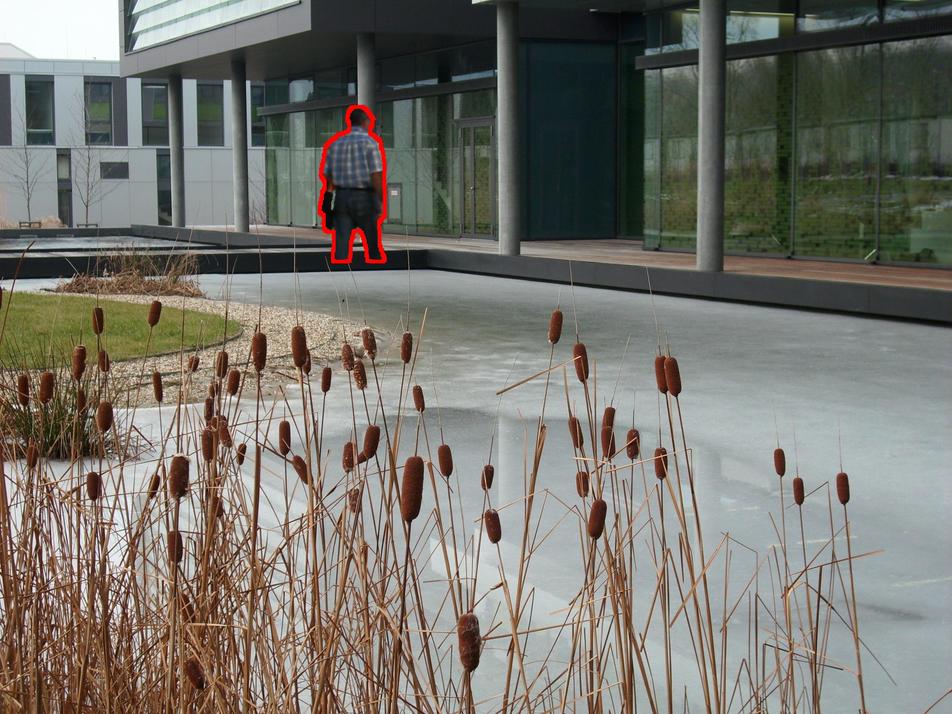} \\
	\includegraphics[width=0.45\linewidth]{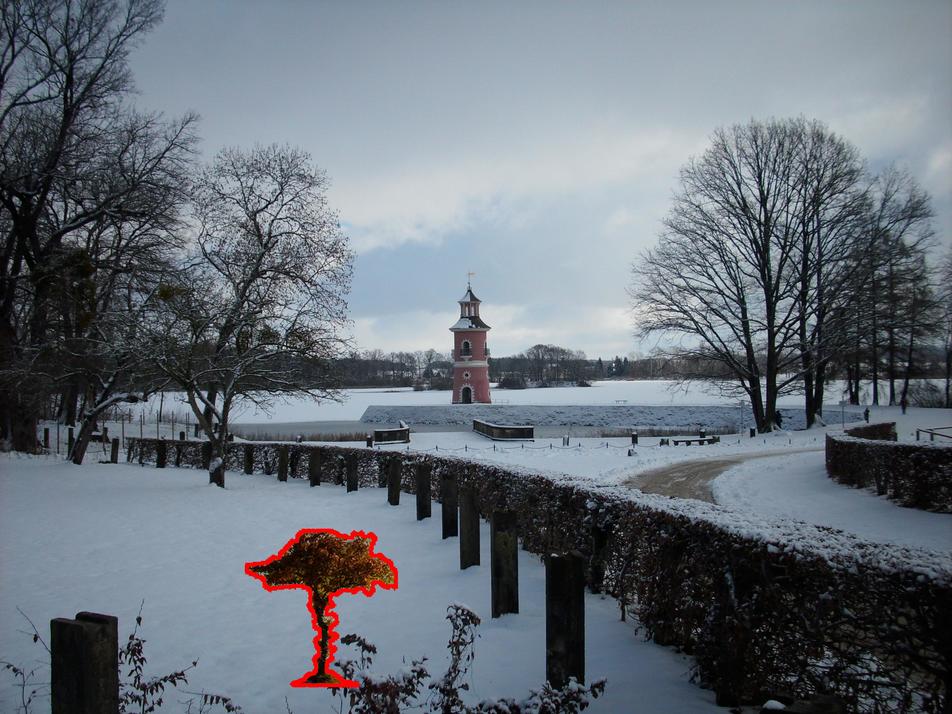} &
	\includegraphics[width=0.45\linewidth]{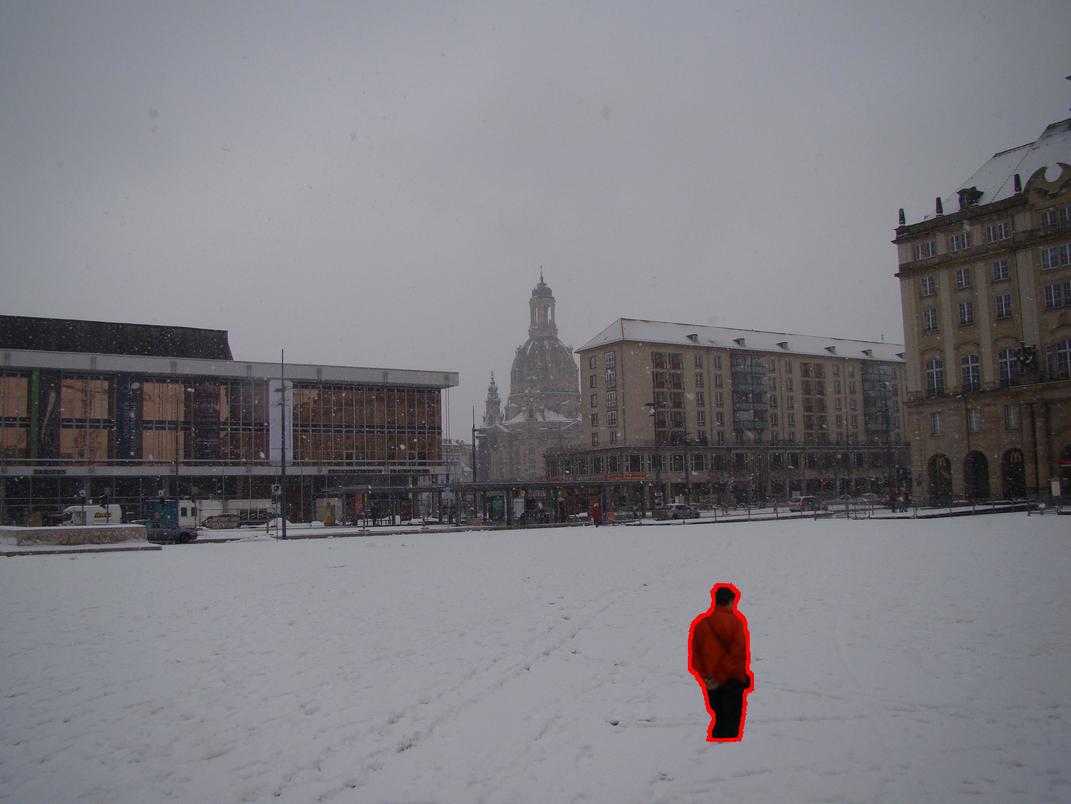} \\
    \end{tabular}
	\caption{Examples from the synthetic Vision/UCID training set. Spliced objects are delimited by a red contour for the sake of clarity.}
	\label{fig:ex_vision_ucid}
\end{figure}

We used all images from 25 devices of the Vision dataset for training, and kept the others for validation, with an approximate 70\%-30\% split, so as to avoid any possible bias.
For each pristine image, we created on the fly a manipulated image
by inserting in a random position one of the UCID objects, selected at random, with random scaling and rotation.
Scaling is such that the size of spliced objects goes from about 1\% to about 10\% of the image size.
Eventually, both pristine and manipulated images are flipped or rotated, and JPEG compressed with QF going from 75 to 100, obtaining a significant augmentation.
Fig.\ref{fig:ex_vision_ucid} shows a few examples of manipulated Vision/UCID images (without rotations).

In the training procedure we used the Adam optimizer with minibatches of 10+10 images and a learning rate of 0.001.
Training took about three days with an Nvidia Tesla P100 GPU.
With the same hardware, testing takes about half a second for a 3072$\times$4096-pixel image, including the noiseprint extraction,
which decreases to 0.01 seconds if the image tiles are already stored in the GPU memory.

\subsection{Preliminary experiments}

The proposed framework aims at the detection of localized manipulations, such as splicing, copy-move, and object removal through content-aware inpainting.
Towards this goal, we instantiated the proposed framework by means of some key design choices.
In particular, we
\begin{itemize}
\item   augmented input RGB bands with the corresponding noiseprint bands;
\item   used Xception \cite{Chollet2017} as feature extractor;
\item   performed aggregation by including all types of pooling;
\item   used two fully connected layers, of size FC1=512 and FC2=256, to perform the final classification.
\end{itemize}

We arrived at these choices as a result of a large number of preliminary experiments, whose description would be dispersive and tedious.
However, we can study experimentally the impact of each individual choice on the performance of the proposed architecture.
To this end, we generated a new dataset, with the same modalities used for the training set, but completely separated from it.
Background images were taken from the Dresden dataset, originally proposed \cite{Gloe2010} for camera model identification,
and manipulated images were created by splicing on them 13 objects taken from the FAU dataset \cite{Christlein2012}
(see again Tab.\ref{tab:datasets} for details on datasets).
After performing the splicing, images were JPEG compressed at high (QF$\geq$95), medium (90$\geq$QF$\geq$85), or low quality (80$\geq$QF$\geq$75),
and eventually resized at scale=0.75 or left unchanged.
Spliced objects can be classified as large, medium, or small,
depending on the largest dimension of their bounding box (after image resizing), set to 1024, 384, or 128, respectively.
Note that, to carry out the large number of tests required by this analysis,
we use a small training set, here, and results indicate main trends but can be improved by a more accurate training.

To assess performance, here and in all subsequent experiments, we classify the whole test set,
compute false positive rate (FPR) and true positive rate (TPR) as a function of the detection threshold, going from 0 to 1,
and obtain the corresponding receiver operating characteristic (ROC) curve.
Eventually, we compute the area under the ROC curve (AUC) as a synthetic measure of performance.

\begin{table}[h]
\centering
\caption{Results of ablation studies on the Dresden/FAU dataset}
\begin{tabular}{|rcl|c|} \hline
   \multicolumn{3}{|c|}{\ru architecture}                            & ~~AUC~~ \\ \hline\hline
   \multicolumn{3}{|c|}{\ru RGB+NP / Xception / all poolings / 512}  &   0.851 \\ \hline
   \ru RGB           & $\to$ & RGB+NP                                &   0.845 \\
   \ru NP            & $\to$ & RGB+NP                                &   0.849 \\ \hline
   \ru Resnet101     & $\to$ & Xception                              &   0.750 \\
   \ru Inception     & $\to$ & Xception                              &   0.745 \\ \hline
   \ru max-pooling   & $\to$ & all poolings                          &   0.800 \\
   \ru avg-pooling   & $\to$ & all poolings                          &   0.808 \\ \hline
   \ru FC1-size  256 & $\to$ & 512                                   &   0.831 \\
   \ru FC1-size 1024 & $\to$ & 512                                   &   0.838 \\ \hline
\end{tabular}
\label{tab:ablation}
\end{table}

\begin{table}[h]
\centering
\caption{Results on subsets from the Dresden/FAU dataset}
\begin{tabular}{|r|c|} \hline
   \ru                sub-dataset & ~~AUC~~ \\ \hline\hline
   \ru                     global &   0.851 \\ \hline
   \ru         large-size objects &   0.855 \\
   \ru        medium-size objects &   0.860 \\
   \ru         small-size objects &   0.875 \\ \hline
   \ru   high-QF JPEG compression &   0.886 \\
   \ru medium-QF JPEG compression &   0.847 \\
   \ru    low-QF JPEG compression &   0.855 \\ \hline
   \ru              original-size &   0.884 \\
   \ru                    resized &   0.841 \\ \hline
\end{tabular}
\label{tab:subdatasets}
\end{table}

In Tab.\ref{tab:ablation} we report the results of our ablation study.
Row 2 refers to the selected architecture,
which uses Xception, takes in input both RGB and noiseprint bands, concatenates vectors given by all pooling types, and use a size-512 FC1 layer.
In all other rows, we modified a single item of this reference architecture.
A number of non-trivial results appear.
First of all, Xception is a much better feature extractor than the two alternatives, Resnet101 \cite{He2016} and InceptionV4 \cite{Szegedy2017}.
We had already observed a similar edge in other applications \cite{Roessler2019} although never so sharp.
The likely reason is Xception's better use of resources, with a much smaller number of parameters to optimize for a given network depth.
It also clearly emerges that using 4 types of pooling together ensures a significant improvement w.r.t. using only one of them.
Using only max-pooling, as suggested by the nature of the problem, is even worse than using average pooling,
probably because of its lower robustness to noise.
As for the size of the first FC layer, 512 appears to be the best choice, although just slightly.
The only controversial choice concerns the input.
In fact, using only the RGB bands or only the noiseprint (NP) bands provides results very close to those of RGB+NP, with a statistically insignificant gap.
Therefore, we refrain from sharp decisions on the input, and will keep testing several options in real-world cases.

We now study the impact of compression, resizing, and splicing size on the performance of the proposed method by collecting results for specific relevant subsets.
A quick look at the numbers of Tab.\ref{tab:subdatasets} makes clear that only minor variations occur across such subsets, with all AUC's in the 0.84--0.89 range.
The largest performance gap is observed between original-size and resized images.
Also JPEG compression affects somewhat the detection performance, although no significant difference emerges between the medium-QF and low-QF cases.
The size of the spliced area, instead, seems to have a minor impact and, contrary to expectation,
relatively small-size splicings are detected more easily that large-size ones.
Note that, on the average, the AUC on specific subsets is larger than the global AUC, but this is a consequence of the higher homogeneity of the tested images.

\subsection{Comparative performance analysis}
\setlength{\tabcolsep}{6pt}
\begin{table*}[!t]
\centering
\caption{Results of all versions of E2E and all references methods on the test datasets.}
\begin{tabular}{|l|c||c|c|c|c|c|c|c|} \hline
   \ru Method                          & supervision & Dresden/FAU  &   ~~DSO-1~~ &  ~~Korus~~ &   ~NC2017~ &   MFC2018  &  MFC2019  & ~average~  \\ \hline\hline
   \ru Xception-resize                 &        weak &       0.609  &     0.539   &     0.527  &     0.513  &     0.570  &    0.516  &     0.546  \\
   \ru Xception-patchwise              &      strong &    \r{0.721} &     0.643   &     0.533  &     0.729  &  \r{0.711} &    0.632  &     0.661  \\ \hline
   \ru CFA \cite{Ferrara2012}          &          -- &       0.507  &     0.584   &  \r{0.598} &     0.593  &     0.539  &    0.526  &     0.558  \\
   \ru DCT \cite{Ye2007}               &          -- &       0.505  &     0.614   &     0.501  &     0.683  &     0.523  &    0.509  &     0.556  \\
   \ru NOI \cite{Mahdian2009}          &          -- &       0.558  &     0.543   &     0.507  &     0.678  &     0.523  & \r{0.726} &     0.589  \\
   \ru NoisePrint \cite{Cozzolino2019} &          -- &       0.611  &  \r{0.821}  &     0.583  &     0.746  &     0.684  &    0.662  &  \r{0.684} \\
   \ru EXIF-SC \cite{Huh2018}          &          -- &       0.599  &     0.721   &     0.496  &     0.709  &     0.670  &    0.655  &     0.642  \\
   \ru SPAM+SVM \cite{Cozzolino2014a}  &        weak &       0.506  &     0.768   &     0.502  &     0.767  &     0.631  &    0.634  &     0.635  \\
   \ru CNN+SVM \cite{Rao2016}          &      strong &       0.593  &     0.728   &     0.568  &  \r{0.798} &     0.702  &    0.679  &     0.678  \\
   \ru LSTM-EnDec \cite{Bappy2019}     &      strong &       0.543  &     0.590   &     0.521  &     0.504  &     0.535  &    0.542  &     0.539  \\ \hline\hline
   \ru E2E-RGB                         &        weak &       0.958  &     0.596   &     0.607  &     0.774  &     0.760  &    0.737  &     0.739  \\
   \ru E2E-NP                          &        weak &       0.874  &  \b{0.924}  &  \b{0.665} &     0.766  &     0.776  &    0.741  &     0.791  \\
   \ru E2E-RGB+NP                      &        weak &       0.914  &     0.790   &     0.619  &     0.762  &     0.765  &    0.765  &     0.769  \\ \hline
   \ru E2E-Fusion                      &        weak &    \b{0.993} &     0.824   &     0.655  &  \b{0.846} &  \b{0.838} & \b{0.787} &  \b{0.824} \\ \hline \end{tabular}
\label{tab:comparison}
\end{table*}

Having justified our design choices,
we now move to compare the performance of the proposed framework with those of suitable baselines and state-of-the-art methods,
using not only our relatively simple Dresden/FAU synthetic dataset,
but also several realistic and challenging datasets widespread in the forensic community.

\subsubsection{Reference methods}

first of all, we consider two natural baselines, both relying on Xception, given its good performance.
The first one, Xception-resize, consists simply in resizing the target image to fit the CNN input, with straightforward training procedure.
Xception-patchwise, instead, works by analyzing the image patch-by-patch, with no resizing and some spatial overlapping, and finally fusing results.
Accordingly, the net is trained to perform binary patch classification.
Since the detector will look for anomalies,
we decided to label only boundary patches as forged, that is, patches including a significant fraction of both background and manipulated areas.
Eventually, the output probabilities are collected in a heatmap,
from which a suitable statistic is extracted (after some tests, we chose the max statistic) and compared with a threshold to make the image-level decision.

Just like our two baselines, methods proposed in the literature can be grouped in two classes.
A few ones work at image level, while the majority work at patch-level, as they pursue forgery localization,
and are converted into image-level detectors through some simple post-processing.

For the first category,
we selected the SPAM+SVM method \cite{Cozzolino2014a}, winner of the First IEEE Forensic Challenge and based on the SPAM steganalytic features \cite{Fridrich2012},
the CNN+SVM method of \cite{Rao2016}, which extract features through a constrained CNN,
and LSTM-EnDec\footnote{Contrary to other supervised methods, we were not able to re-train the network on our data and used the original model in experiments.} \cite{Bappy2019},
which uses a long-short term memory recurrent neural network to detect pristine/forged spatial transitions.
For the second category, we consider several forgery localization methods converted into image-level detectors.
In particular, we selected the best performing methods resulting from the analysis carried out in \cite{Cozzolino2019}, that is,
CFA \cite{Ferrara2012}, which exploits features related to the color-filter array,
DCT \cite{Ye2007}, based on the analysis of double-quantized DCT coefficients,
NOI \cite{Mahdian2009}, looking for spatial inconsistencies in the noise level,
EXIF-SC \cite{Huh2018}, looking for anomalies in the image leveraging the EXIF metadata during the training phase,
and Noiseprint \cite{Cozzolino2019}, which extracts and analyzes an image fingerprint where camera model-related artifacts are emphasized.
All these methods compute a heatmap representing the probability that a certain patch has been manipulated.
To make the image-level decision we extract several statistics from such heatmaps: mean, maximum, and $q$-quantile, with $q \in \{5,10,\ldots,95\}$,
selecting the best one in terms of AUC performance separately for each method.
Note that all these latter methods are blind, that is, they require no training on forged images or patches.

\subsubsection{Datasets}

\begin{figure}[t]
	\centering
	\setlength{\tabcolsep}{1mm}
	\begin{tabular}{cc}
	\includegraphics[width=0.45\linewidth, height=3cm]{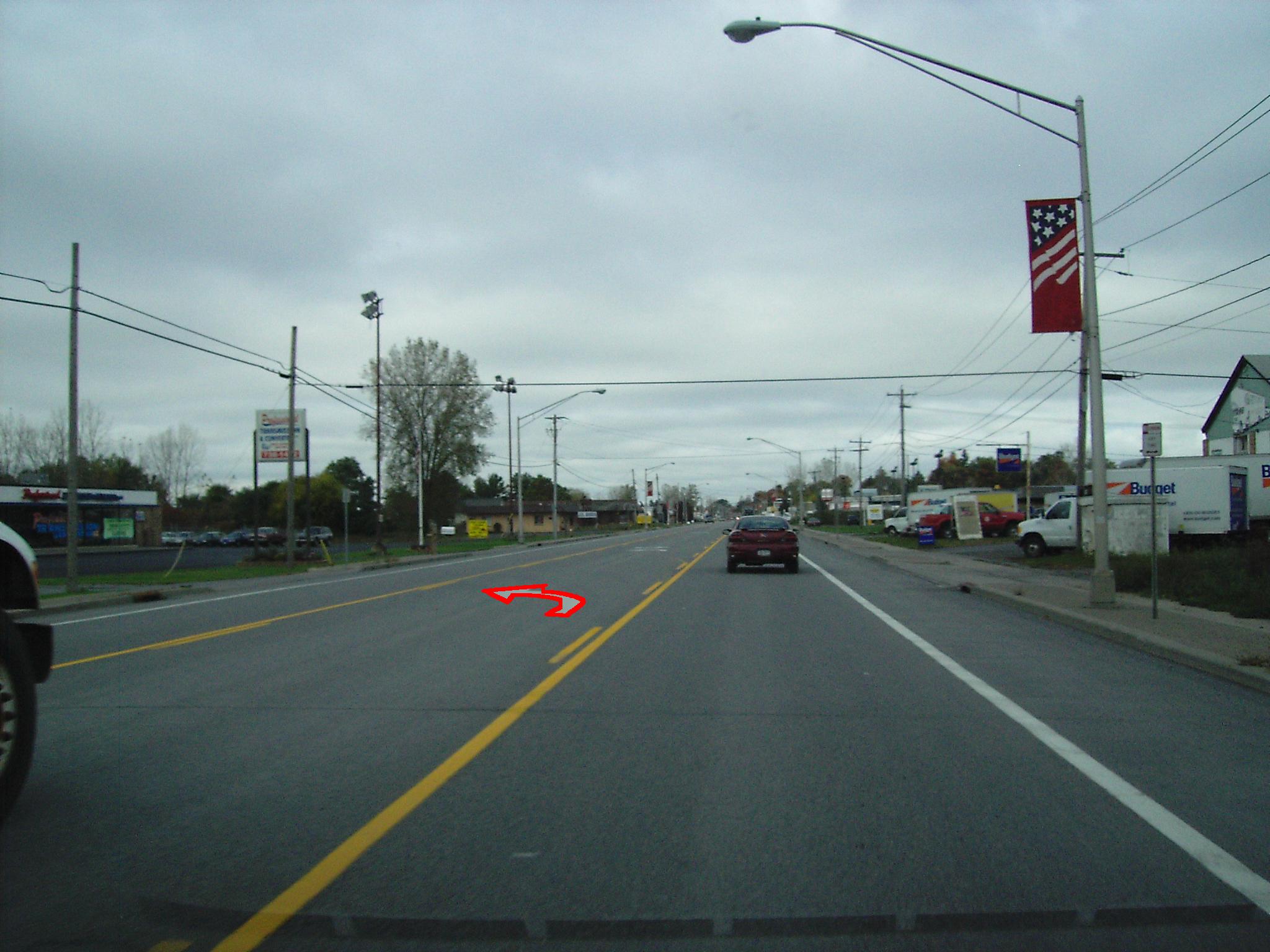} &
	\includegraphics[width=0.45\linewidth, height=3cm]{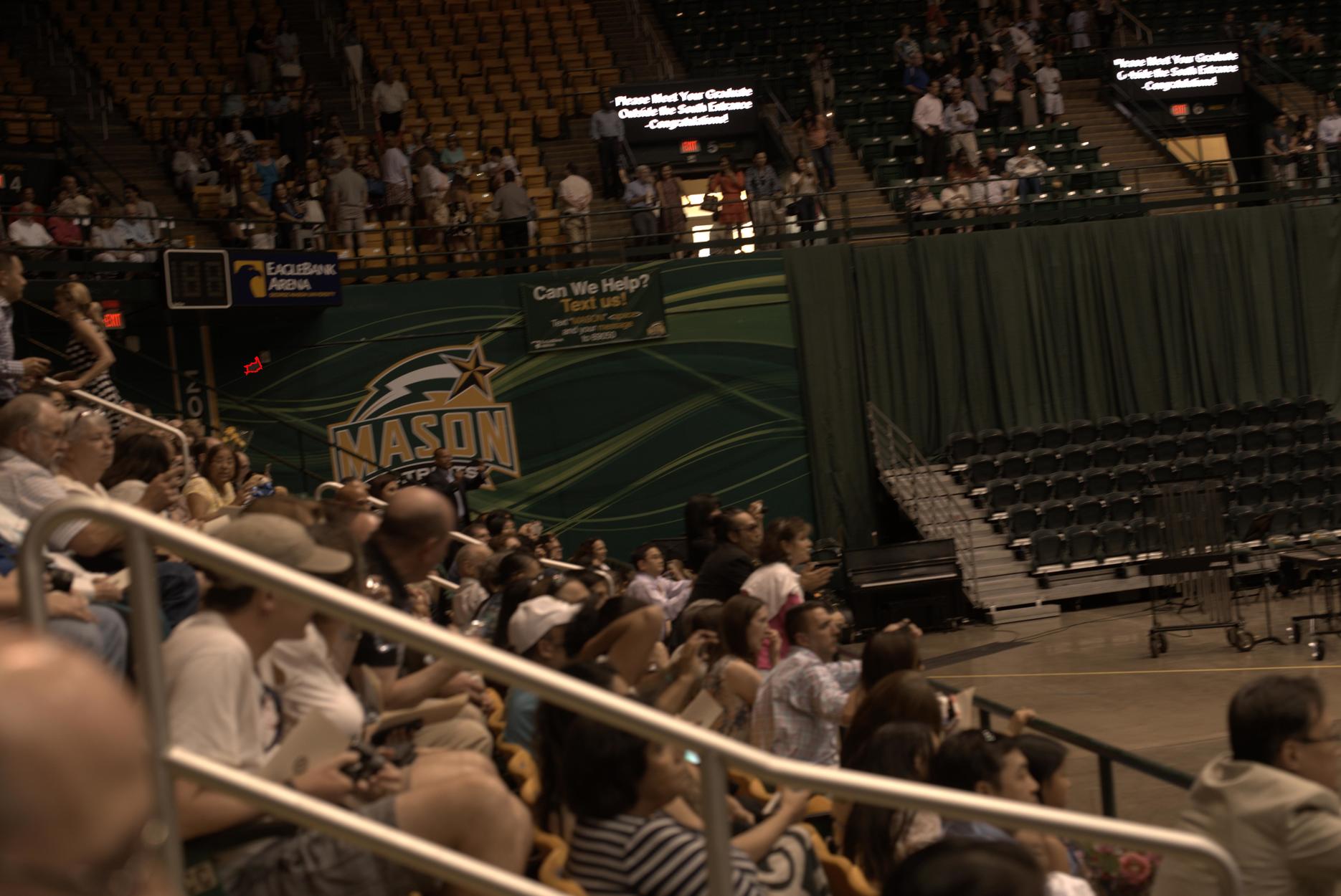} \\
	\includegraphics[width=0.45\linewidth, height=3cm]{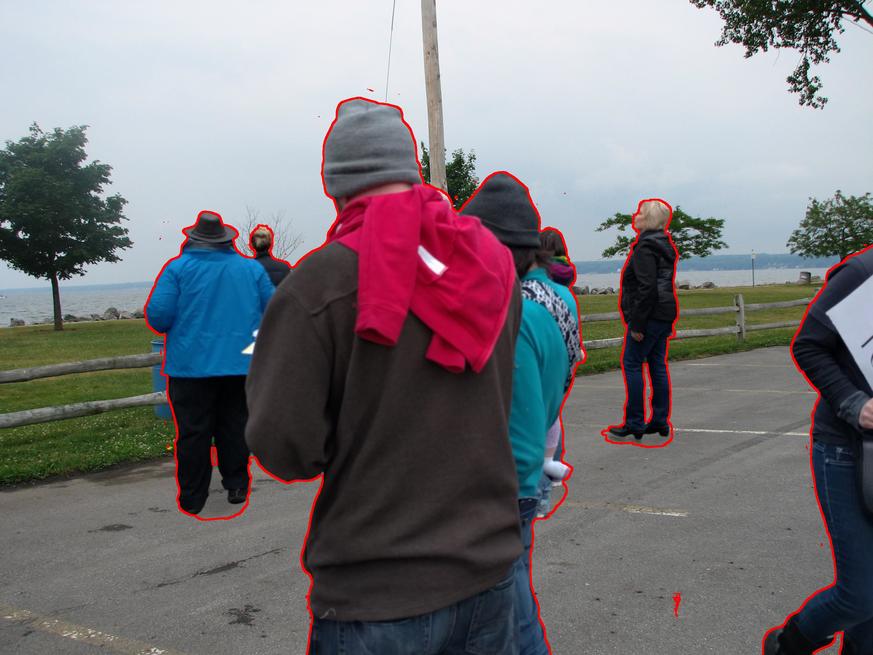} &
	\includegraphics[width=0.45\linewidth, height=3cm]{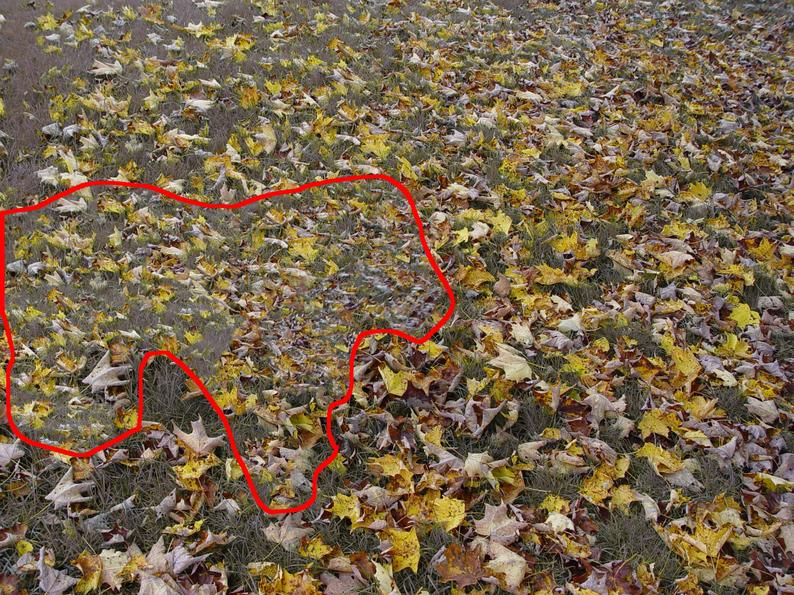} \\
    \end{tabular}
	\caption{Examples from the NIST datasets.}
	\label{fig:ex_medifor}
\end{figure}

for performance assessment, besides our synthetic Dresden/FAU dataset,
we consider several more datasets, widely used in the forensics community, with markedly different characteristics.
DSO-1 \cite{Carvalho2013} features only splicings, with little or no post-processing.
In Korus \cite{Korus2016a}, instead, both splicings and copy-moves are present.
Both datasets include only large-size high-quality images, not even compressed in the case of Korus.
A very different, and much more challenging, scenario is given by the NC2017, MFC2018, and the very recent MFC2019 datasets \cite{Guan2019},
developed by NIST\footnote{https://www.nist.gov/itl/iad/mig/media-forensics-challenge-2019-0} in the context of the Medifor initiative.
Images of these datasets have been manually doctored, often with multiple and possibly overlapping manipulations of various types.
In addition, they have wildly different sizes and quality levels, and have been subject to several anti-forensics measures to prevent easy detection and localization of forgeries.
For our tests, we kept all images with splicing, copy-move, inpainting, or computer-generated material.
The reader is referred to Tab.\ref{tab:datasets} and to the original papers for more details,
while some example images are shown in Fig.\ref{fig:ex_medifor}

\subsubsection{Numerical results}

\begin{figure*}[t]
	\centering
	\begin{tabular}{cc}
		\includegraphics[width=0.48\linewidth]{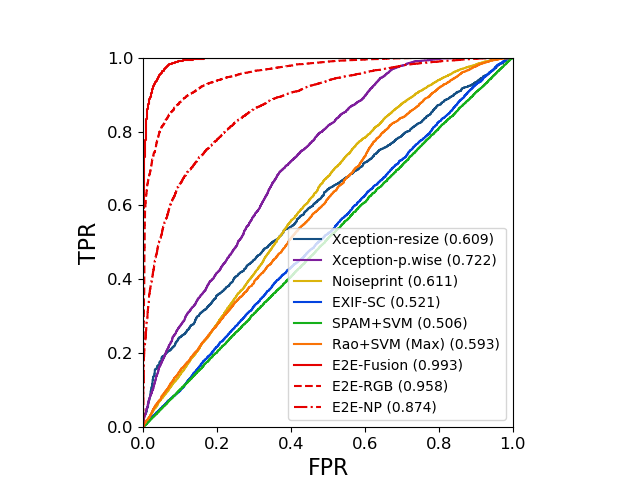} &
		\includegraphics[width=0.48\linewidth]{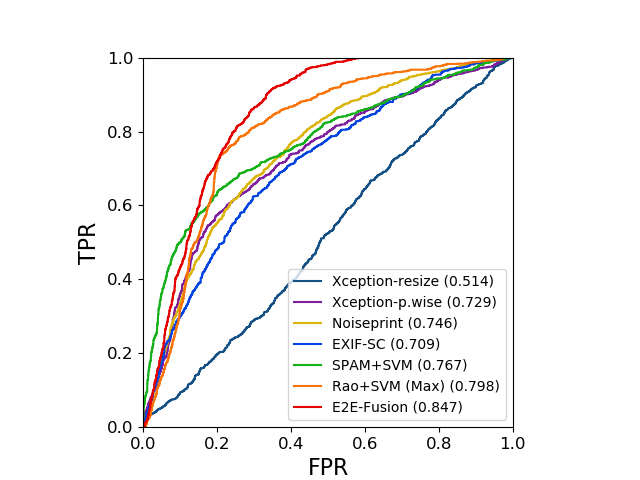}  \\
		\includegraphics[width=0.48\linewidth]{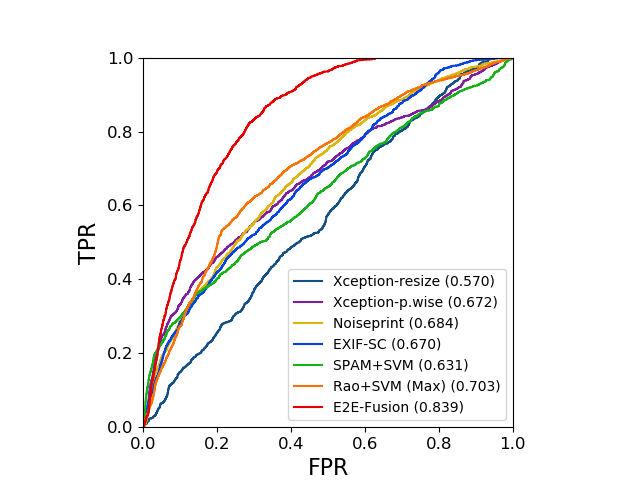} &
		\includegraphics[width=0.48\linewidth]{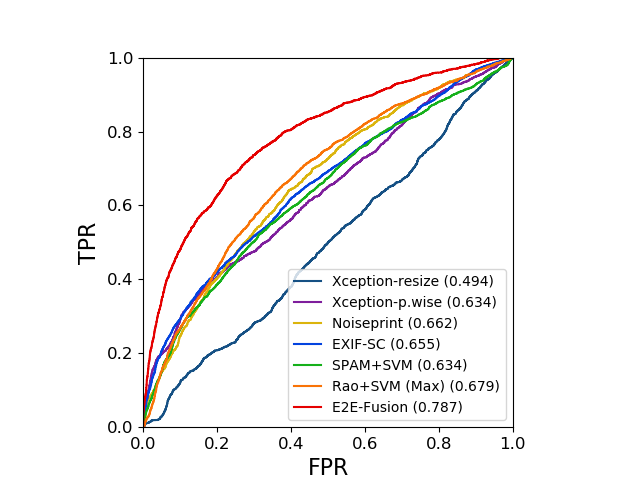} \\
	\end{tabular}
	\caption{ROC curves on Dresden/FAU (top-left), NC2017, MFC2018, MFC2019 (bottom-right) datasets.
             For the sake of clarity, ROCs are shown only for selected methods:
             the best proposed (E2E-Fusion), the two baselines, and the best references (SPAM+SVM, CNN+SVM, Noiseprint, EXIF-SC).
             Only for Dresden/FAU we also show other E2E versions.
             E2E-Fusion is always clearly, and almost uniformly, the best.
             The resizing-based baseline always the worst.}
	\label{fig:dataset_ROC}
\end{figure*}

in Tab.\ref{tab:comparison} we report the detection AUC for all reference and proposed methods on all test datasets.
Next to each method, in column 2, we give the level of supervision it requires, strong (pixel-wise ground truth), weak (only image label), or -- (none) for blind methods.
In the upper part of the table we group all reference methods, including our two baselines, and in the lower part all version of the proposed method with end-to-end (E2E) training.
Best results are highlighted in red for reference methods and in blue for our proposal.
In Fig.\ref{fig:dataset_ROC} we also show ROC curves for a subset of methods (for readability) and datasets (for space) characterized by very different features.
On the Dresden/FAU dataset, disjoint from the training Vision/UCID dataset, but well-aligned with it,
the proposed method (E2E-RGB+NP) largely outperforms all references,
with a gain of almost 20 percent points over the best one, the strongly supervised Xception-patchwise.
Guided by the outcomes of preliminary experiments,
together with the ``best'' version, with RGB+NP input, we consider also the versions with only RGB and only NP inputs.
To our surprise, E2E-RGB provides a further significant performance improvement.
Our explanation for this phenomenon is the strong eterogeneity of the input:
since RGB bands and noiseprints have quite different statistics, the net may have a hard time processing them jointly.
To confirm such hypothesis, we considered a further versions of the proposed method,
where the networks trained on RGB-only, NP-only, and RGB+NP inputs are fused afterwards by a trivial average of scores.
This strategy proved successful, with the new version, E2E-Fusion,
providing almost perfect detection (see also the top-left ROC in Fig.\ref{fig:dataset_ROC}), thus confirming our conjecture.

Moving to the DSO-1 dataset, we observe again a large gain, more than 10 percent points, of the best E2E method over the best reference.
On this dataset, Noiseprint provides an especially good performance.
a phenomenon already observed in \cite{Cozzolino2019}, and likely related to all images being JPEG compressed at high-quality.
Accordingly, also E2E works best with only noiseprints as input, with no fusion.
Images of the Korus dataset, instead, are uncompressed.
This removes a major source of forensic traces, which impacts all methods, some of which exhibit a 0.5 AUC, equivalent to coin tossing.
CFA (relying on color filter array properties) and Noiseprint, keep providing decent results,
however they trail all E2E versions, featuring AUC's between 0.60 and 0.66.

Turning to the more challenging NIST datasets, the general behavior does not change, with E2E working generally better than reference methods.
The best reference method is not always the same for all such datasets: CNN+SVM for NC2017, Xception-patchwise for MFC2018, NOI for MFC2019.
On the contrary, E2E-Fusion is always the best version of proposed method, and the best overall,
with a significant performance gain over the best reference, going from 0.048 (NC2017) to 0.127 (MFC2018).

The final column shows the average over all datasets, which confirms all above observations.
We only underline, in passing, that the Xception-resize baseline, as expected, performs quite poorly due to the loss of precious high-frequency details,
while the Xception-patchwise baseline is among the best references, although it is fair to recall that it requires strong supervision.

\begin{figure}[t]
	\centering
	\begin{tabular}{c}
		\includegraphics[width=0.80\linewidth]{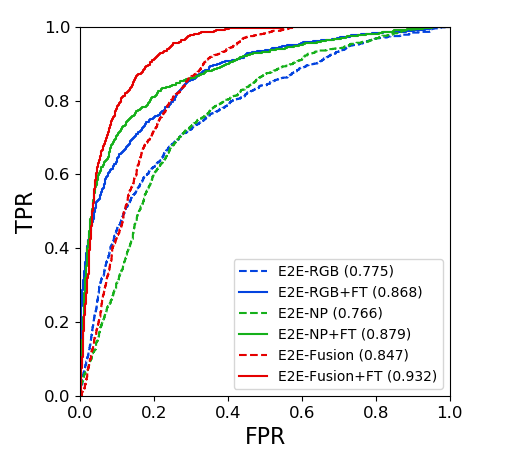} \\
		\includegraphics[width=0.80\linewidth]{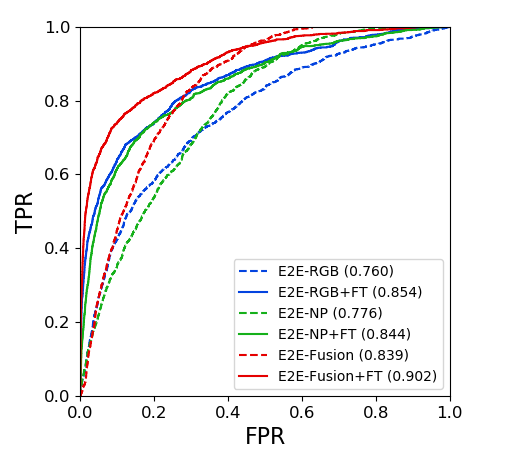}
	\end{tabular}
	\caption{ROC curves of all E2E variants on NC2017 (top) and MFC2018 (bottom) without (dashed lines) and with (solid) fine-tuning on the NIST development sets.
    Fine-tuning provides a significant gain in all cases.}
	\label{fig:NIST_ROC_FineTuned}
\end{figure}

A general observation is that the performance of E2E is consistently good in all cases (with a small dip on Korus),
including the NIST datasets, despite their great variety and the abundance of counter-forensic measures.
This is all the more remarkable, considering that the network was trained on a dataset, Vision/UCID, lacking such a diversity.
Therefore, we carried out a further experiment on NC2017 and MFC2018,
in which the E2E methods are fine-tuned on their respective development sets, provided by NIST together with the test sets.
Results are reported in Tab.\ref{tab:NC2017-finetune} and Tab.\ref{tab:MFC2018-finetune}, while Fig.\ref{fig:NIST_ROC_FineTuned} shows the corresponding ROC curves.
It is clear that fine-tuning on the development set, certainly more aligned with the test set than Vision/UCID, grants further performance gains.
Over the whole dataset (``all'' column) the best AUC, obtained always with E2E-Fusion, grows from 0.846 to 0.932 on NC2017, and from 0.838 to 0.902 on MFC2018.
The larger improvement on NC2017 can be attributed to better development-test alignment and lighter counter-forensic actions.
In any case, results are extremely satisfactory for such challenging datasets.

In the tables, taking advantage of the auxiliary information provided with these datasets, we also provide analytic results for each type of forgery.
Although E2E is trained only on splicing, it works well also on all other localized manipulations.
The most interesting phenomenon we could spot from these data is the performance drop on computer-generated fakes.
In NC2017 the AUC for these manipulations was very high, above 0.93 without fine-tuning, lowering dramatically (0.80) in MFC2018.
Probably, this is the effect of the fast pace of progress in the quality of such manipulations.

\setlength{\tabcolsep}{3pt}
\begin{table}
\centering
\caption{Results of E2E methods on NC2017 w/o and with finetuning.}
\begin{tabular}{|l|c||c|c|c|c|c|} \hline
   \ru Method           & f.t. &  ~~all~~  & splicing  &   ~~CM~~  & inpaint.  &   ~~CG~~  \\ \hline\hline
   \ru E2E-RGB          &      &    0.774  &    0.829  &    0.819  &    0.694  & \r{0.949} \\
   \ru E2E-NP           &      &    0.765  &    0.774  &    0.752  &    0.762  &    0.902  \\
   \ru E2E-RGB+NP       &  n   &    0.762  &    0.816  &    0.832  &    0.693  &    0.921  \\
   \ru E2E-Fusion       &      & \r{0.846} & \r{0.860} & \r{0.870} & \r{0.809} &    0.932  \\ \hline\hline
   \ru E2E-RGB          &      &    0.868  &    0.871  &    0.887  &    0.833  & \b{0.937} \\
   \ru E2E-NP           &      &    0.879  &    0.799  &    0.849  &    0.914  &    0.880  \\
   \ru E2E-RGB+NP       &  y   &    0.913  &    0.837  &    0.893  &    0.939  &    0.885  \\
   \ru E2E-Fusion       &      & \b{0.932} & \b{0.884} & \b{0.911} & \b{0.950} &    0.935  \\ \hline
\end{tabular}
\label{tab:NC2017-finetune}
\end{table}

\setlength{\tabcolsep}{3pt}
\begin{table}
\centering
\caption{Results of E2E methods on MFC2018 w/o and with finetuning.}
\begin{tabular}{|l|c||c|c|c|c|c|} \hline
   \ru Method           & f.t. &  ~~all~~  & splicing  &   ~~CM~~  & inpaint.  &   ~~CG~~  \\ \hline\hline
   \ru E2E-RGB          &      &    0.760  &    0.808  &    0.705  &    0.696  &    0.730  \\
   \ru E2E-NP           &      &    0.775  &    0.805  &    0.750  &    0.744  & \r{0.817} \\
   \ru E2E-RGB+NP       &  n   &    0.765  &    0.795  &    0.733  &    0.734  &    0.786  \\
   \ru E2E-Fusion       &      & \r{0.838} & \r{0.860} & \r{0.811} & \r{0.811} &    0.799  \\ \hline\hline
   \ru E2E-RGB          &      &    0.854  &    0.874  &    0.823  &    0.802  &    0.851  \\
   \ru E2E-NP           &      &    0.844  &    0.868  &    0.819  &    0.811  &    0.864  \\
   \ru E2E-RGB+NP       &  y   &    0.867  &    0.893  &    0.842  &    0.824  &    0.874  \\
   \ru E2E-Fusion       &      & \b{0.902} & \b{0.925} & \b{0.877} & \b{0.856} & \b{0.910} \\ \hline
\end{tabular}
\label{tab:MFC2018-finetune}
\end{table}

\subsection{Towards forgery localization}

The E2E framework was conceived and trained with the goal of making global decisions, leaving the problem of localization to other tools.
However, just like localization tools can be used for detection through suitable fusion, the proposed detection framework can be recast to provide also some localization information.
In the following subsections we provide some insight into how the system exploits and combines local information coming from all over the image,
and how this can be exploited towards forgery localization.

\subsubsection{Activation Maps}
\begin{figure*}[t]
	\centering
	\setlength{\tabcolsep}{1mm}
	\begin{tabular}{cccc}
    \includegraphics[width=0.23\linewidth]{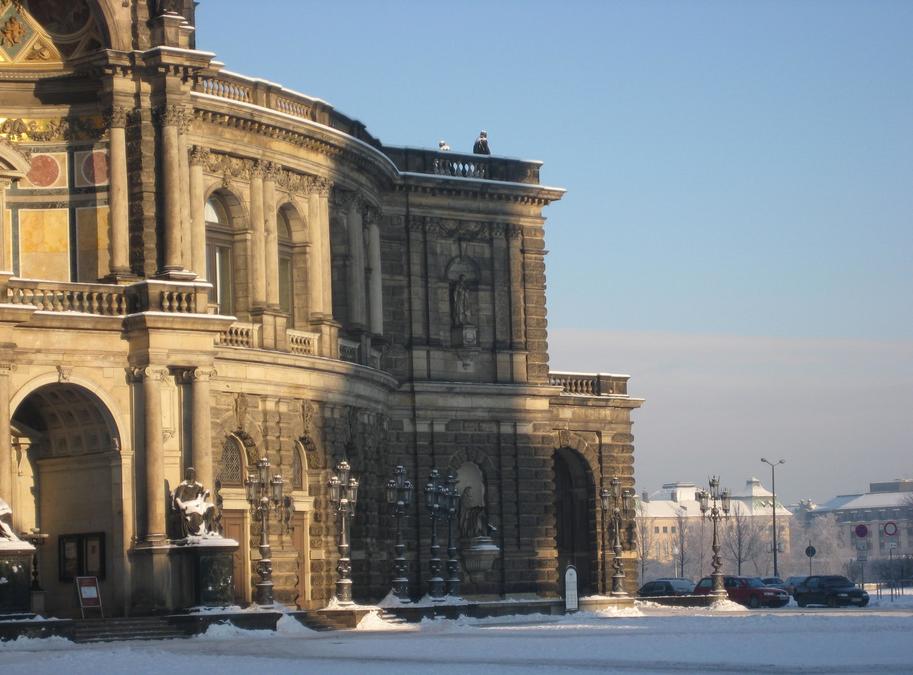}        &
    \includegraphics[width=0.23\linewidth]{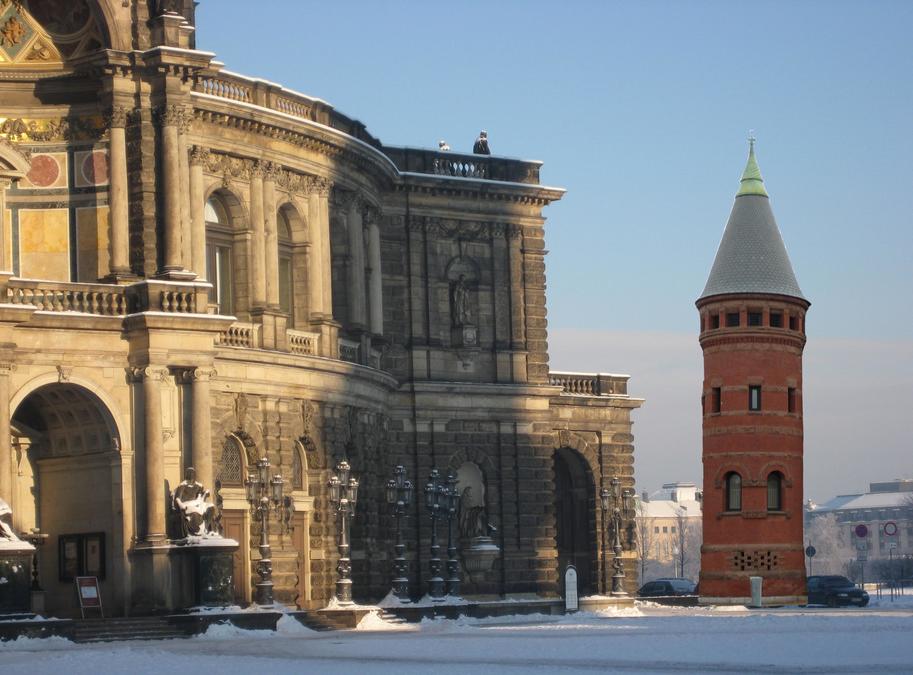}          &
    \includegraphics[width=0.23\linewidth]{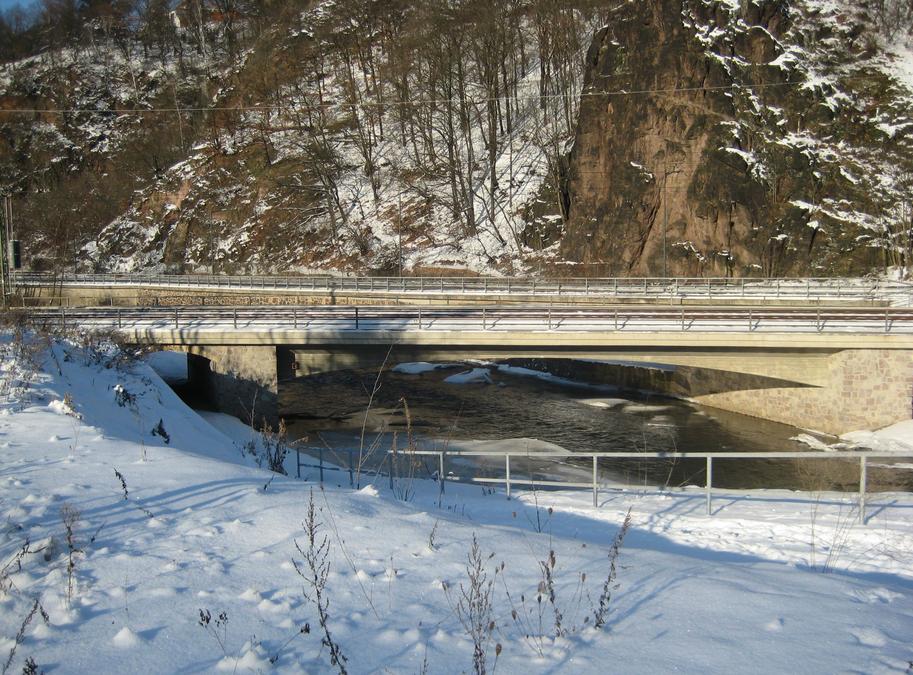}        &
    \includegraphics[width=0.23\linewidth]{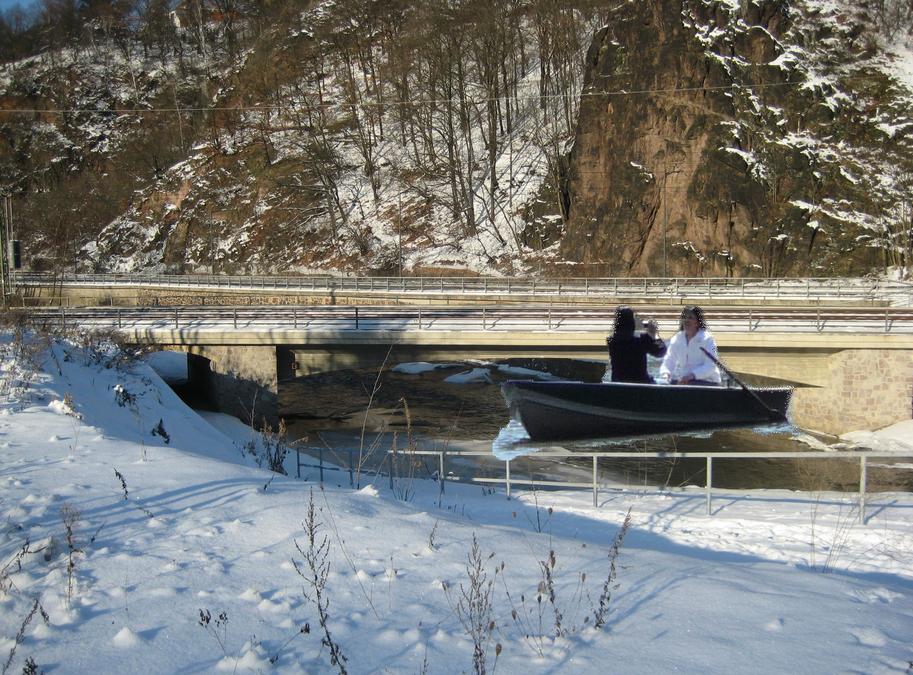}          \\
    \includegraphics[width=0.23\linewidth]{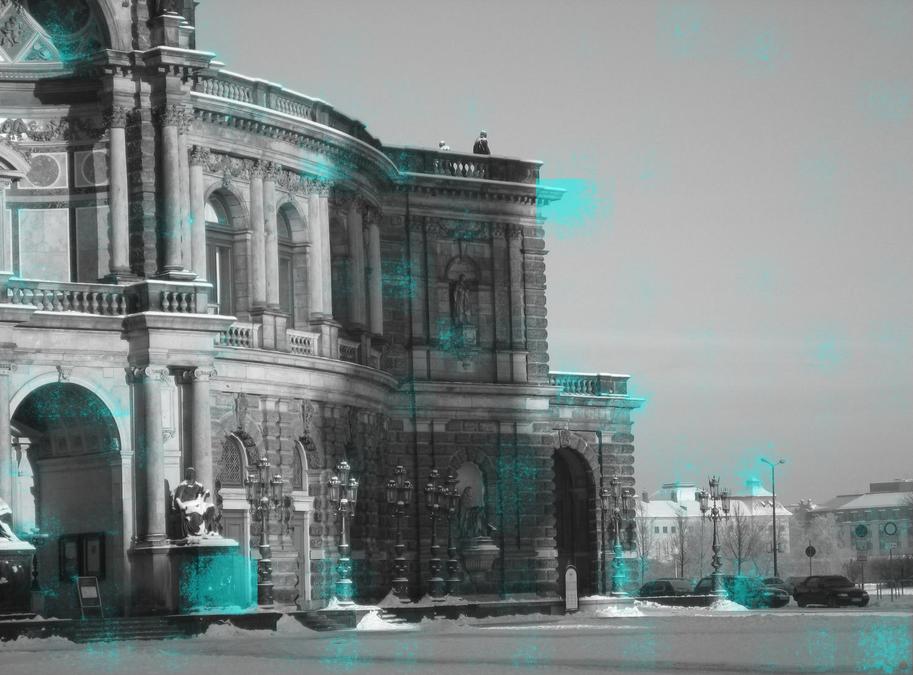} &
    \includegraphics[width=0.23\linewidth]{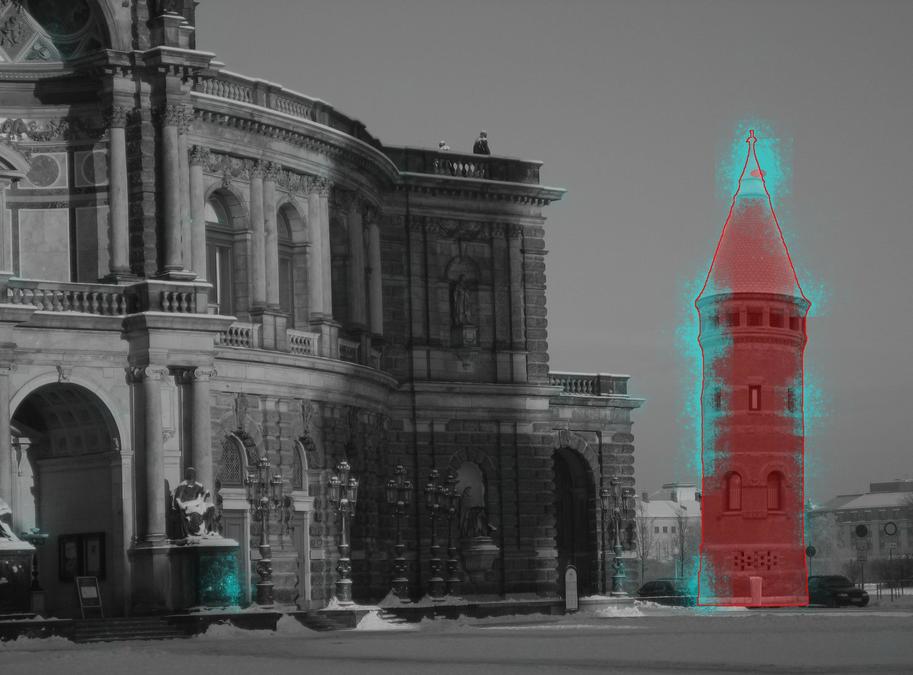}   &
    \includegraphics[width=0.23\linewidth]{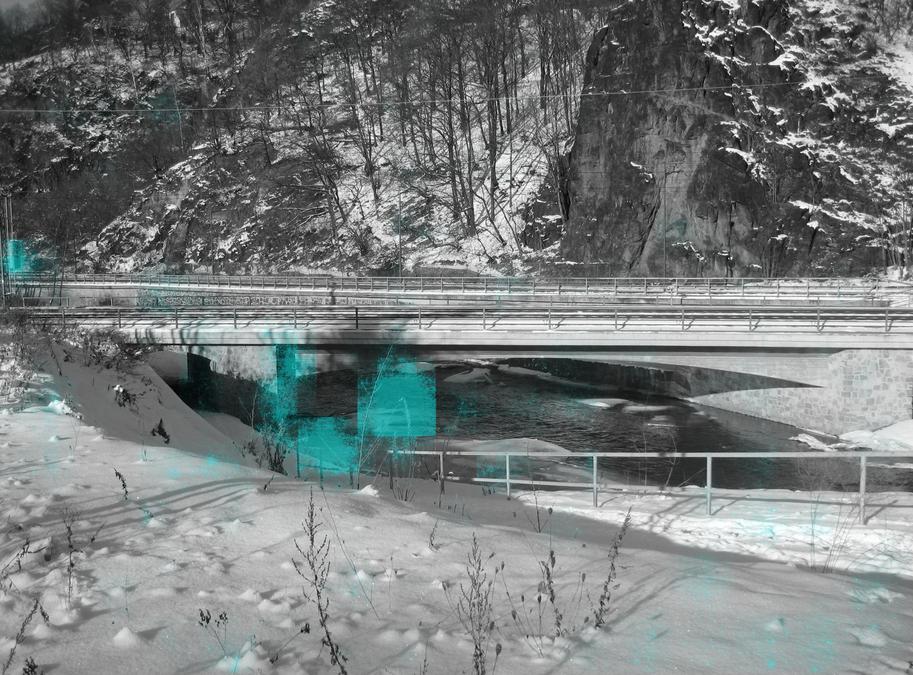} &
    \includegraphics[width=0.23\linewidth]{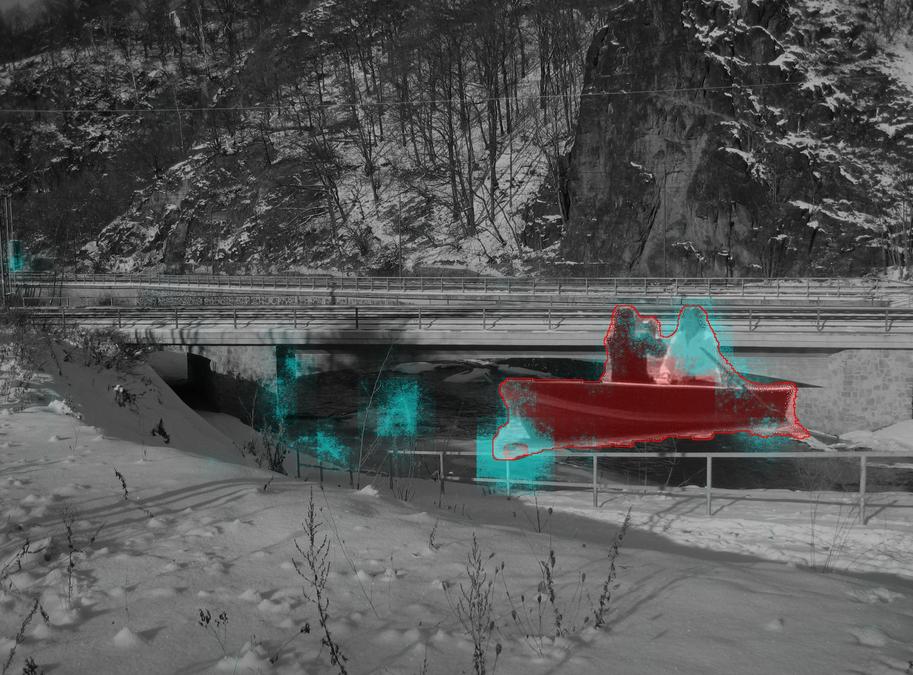}   \\
    \end{tabular}
    \caption{Example images (top) and activation maps (bottom) from the Dresden/FAU dataset.
    Pristine images are on the odd columns, forged images (with hand-made splicings for higher visibility) on the even columns.
    Active patches are superimposed in cyan to the gray-scale/red-scale version of the images.}
    \label{fig:dresden_activations}
\end{figure*}

first of all, we try to investigate the impact of each patch of the image on the final decision.
To this end, we consider a simplified framework in which only the max pooling is used.
Given this hard selection rule, we can easily compute a spatial activation map which counts how many features each patch contributes to the overall feature vector.
Such a map, however, would be extremely coarse, due the low resolution of patch-wise analysis.
Therefore, we combine it with the a high-resolution map,
the Grad-CAM (guided gradient weighted class activation map) obtained by backpropagating the loss gradient to the full-resolution input \cite{Selvaraju2017}.
In Fig.\ref{fig:dresden_activations} we show some results for images of the Dresden/FAU dataset (hand-made to look more realistic).
For this synthetic dataset, we have the pristine version of each manipulated image, so we can analyze the network behavior in both circumstances.
In all cases, the network focuses on high-activity regions, often corresponding to object boundaries.
When there is no manipulation, the salient regions are scattered all over the image.
On the contrary, when a splicing takes place, they tend to concentrate on the boundaries of the spliced object, proving that the system has learned to look at these patches to make its decisions.
Therefore, when a forged image is detected, this activation provides hints about the possible site of the manipulation.

\subsubsection{ROI-based Analysis}
moving towards forgery localization, we can obtain some interesting results by leveraging the flexibility of the proposed framework.
Indeed, since the system can analyze images of any size, it can also analyze regions of interest (ROI) selected by the user based on the previous activation map or any other criterion.
If the ROI contains manipulated material, the system will likely provide a large probability of manipulation (score, from now on).
Therefore, the system can be used in supervised modality to test suspicious objects.
Also, it can be recast to perform automatic box-like localization.
In fact, once features have been computed and stored for all patches, the aggregation and classification phases are extremely simple, with light-speed processing.
Therefore, one can easily test a large number of boxes and select automatically as ROI those with the largest scores,
obtaining a rough but effective form of localization.

\begin{figure*}[t]
	\centering
	\setlength{\tabcolsep}{1mm}
	\begin{tabular}{cccc}
    \includegraphics[width=0.240\linewidth]{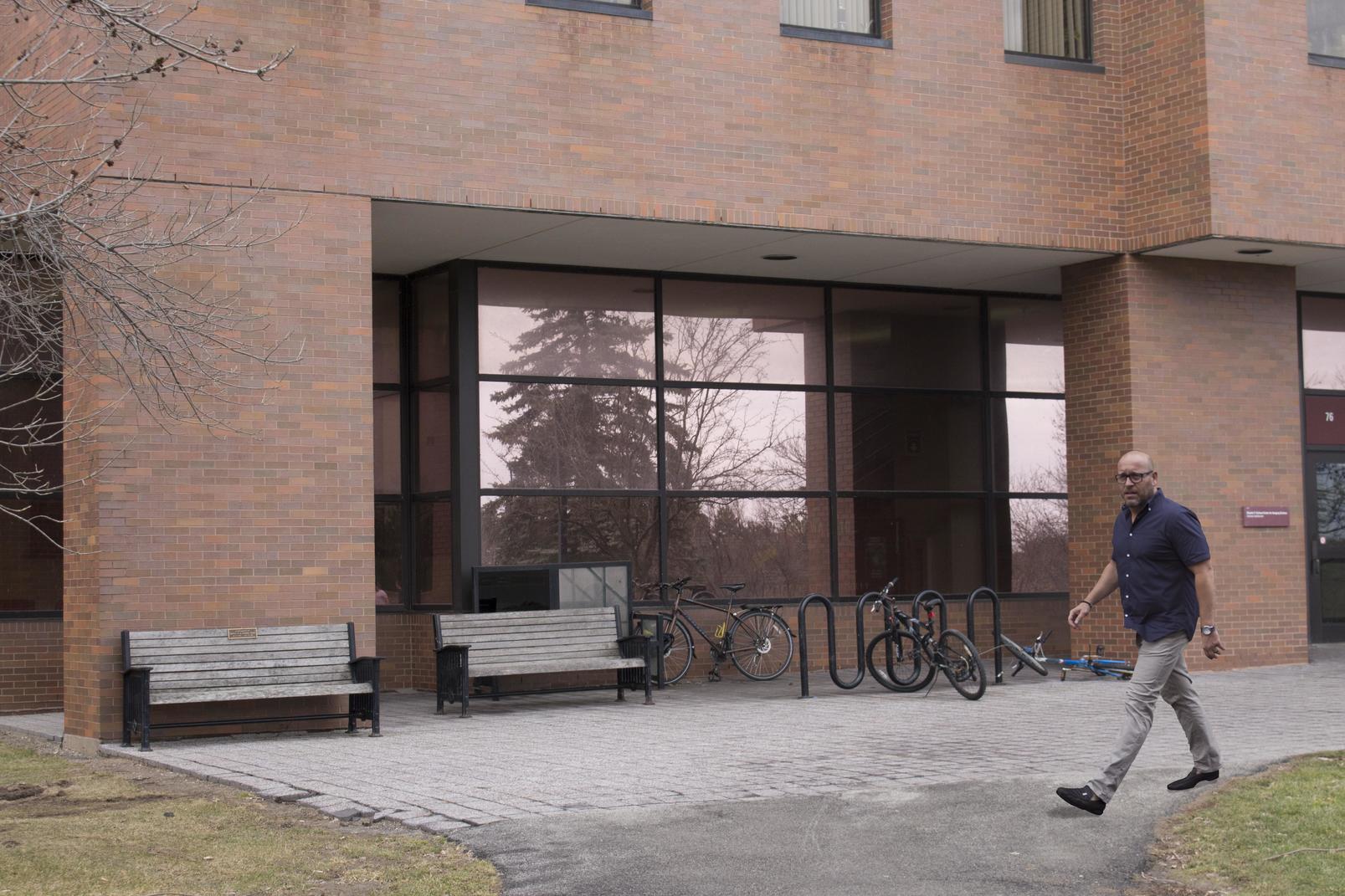}         &
    \includegraphics[width=0.240\linewidth]{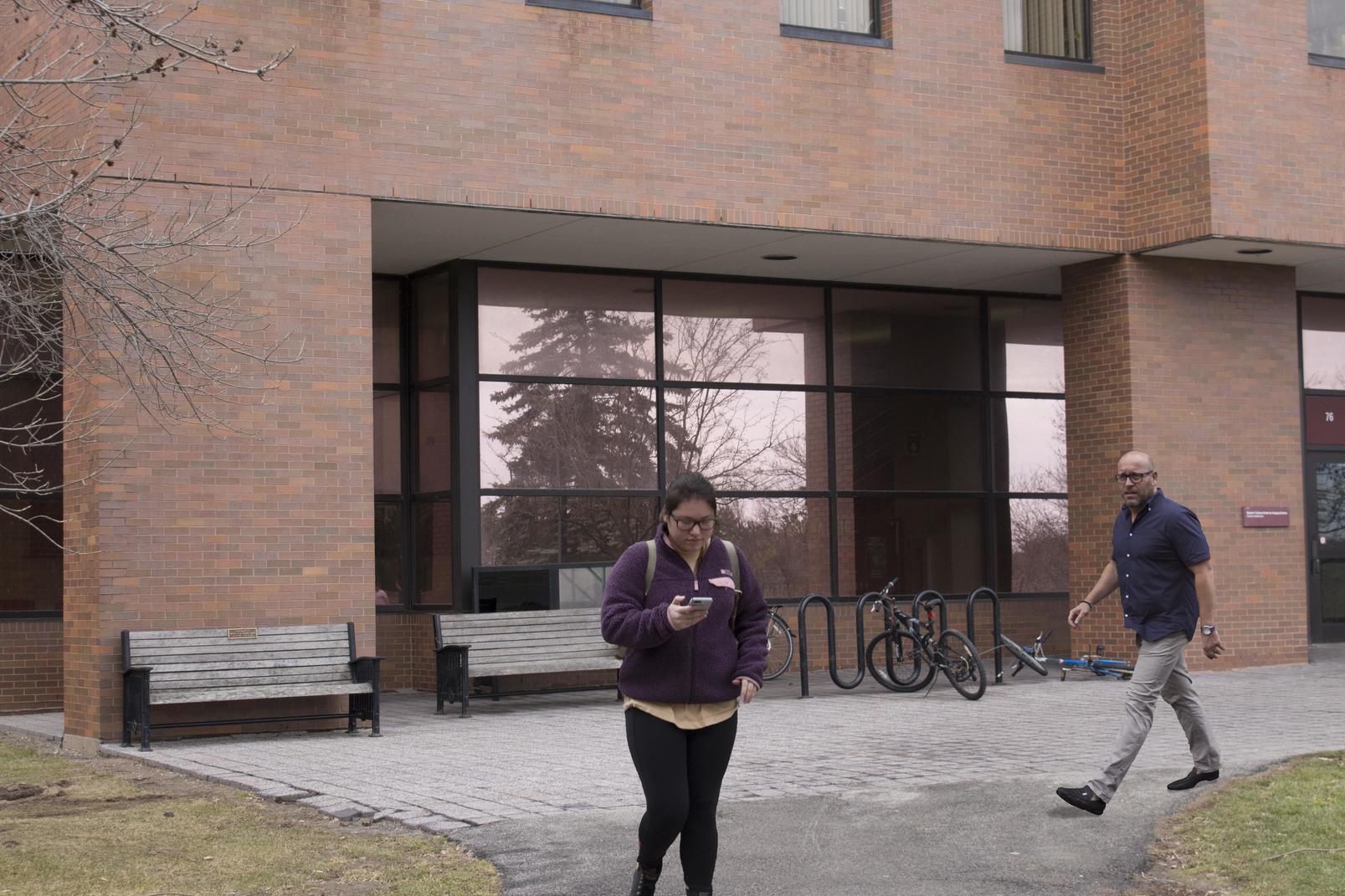}         &
    \includegraphics[width=0.240\linewidth]{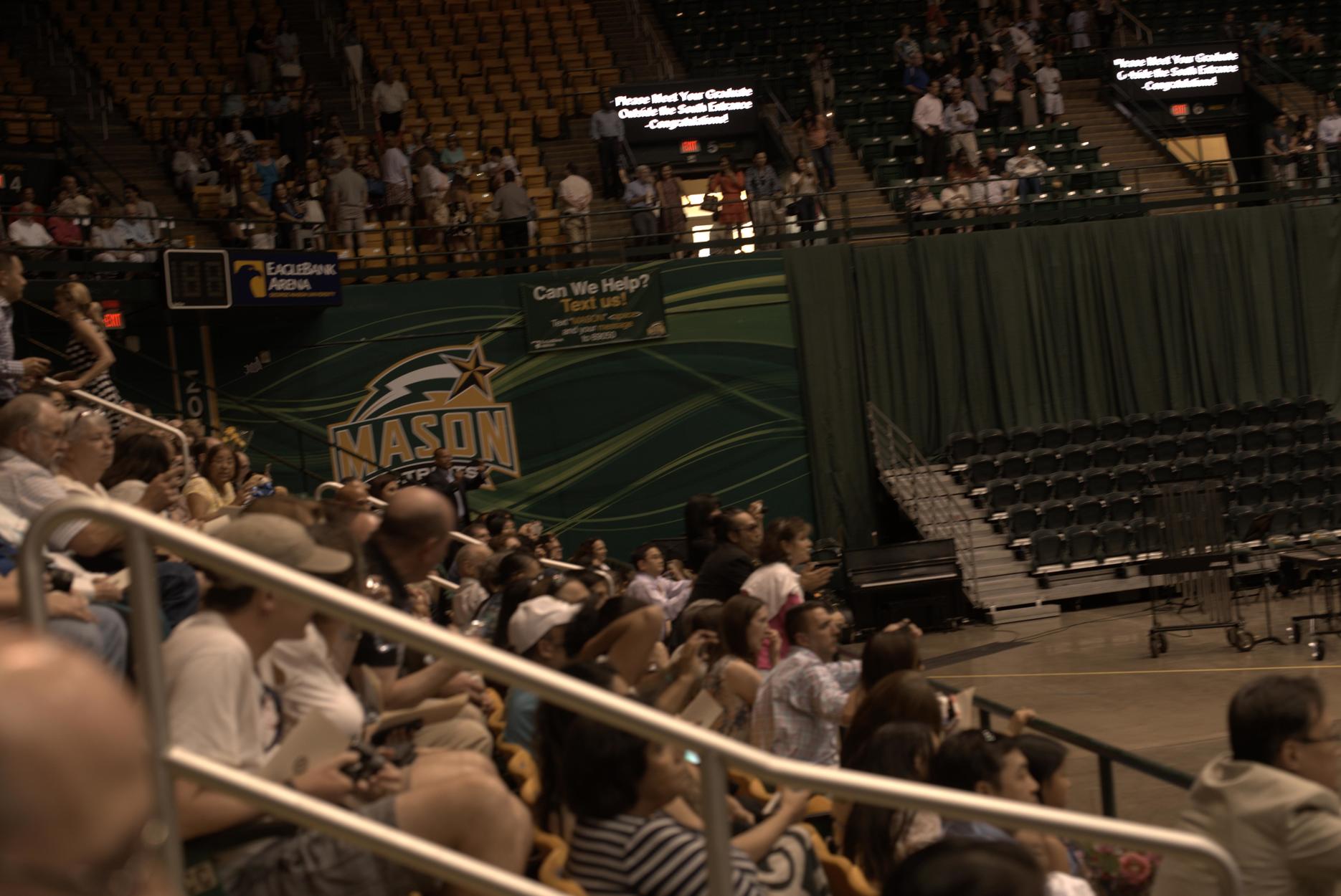}         &
    \includegraphics[width=0.214\linewidth]{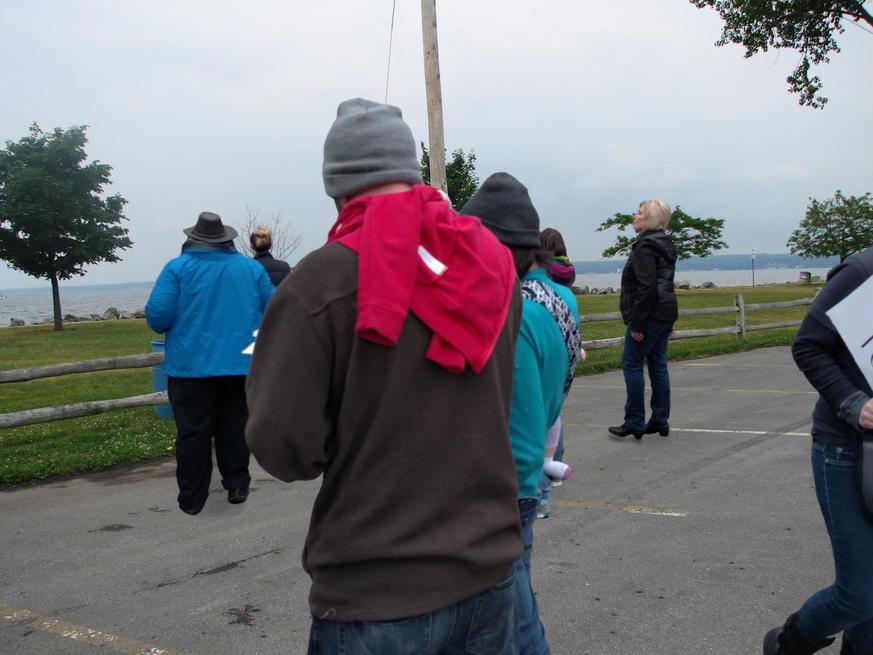}         \\
    \includegraphics[width=0.240\linewidth]{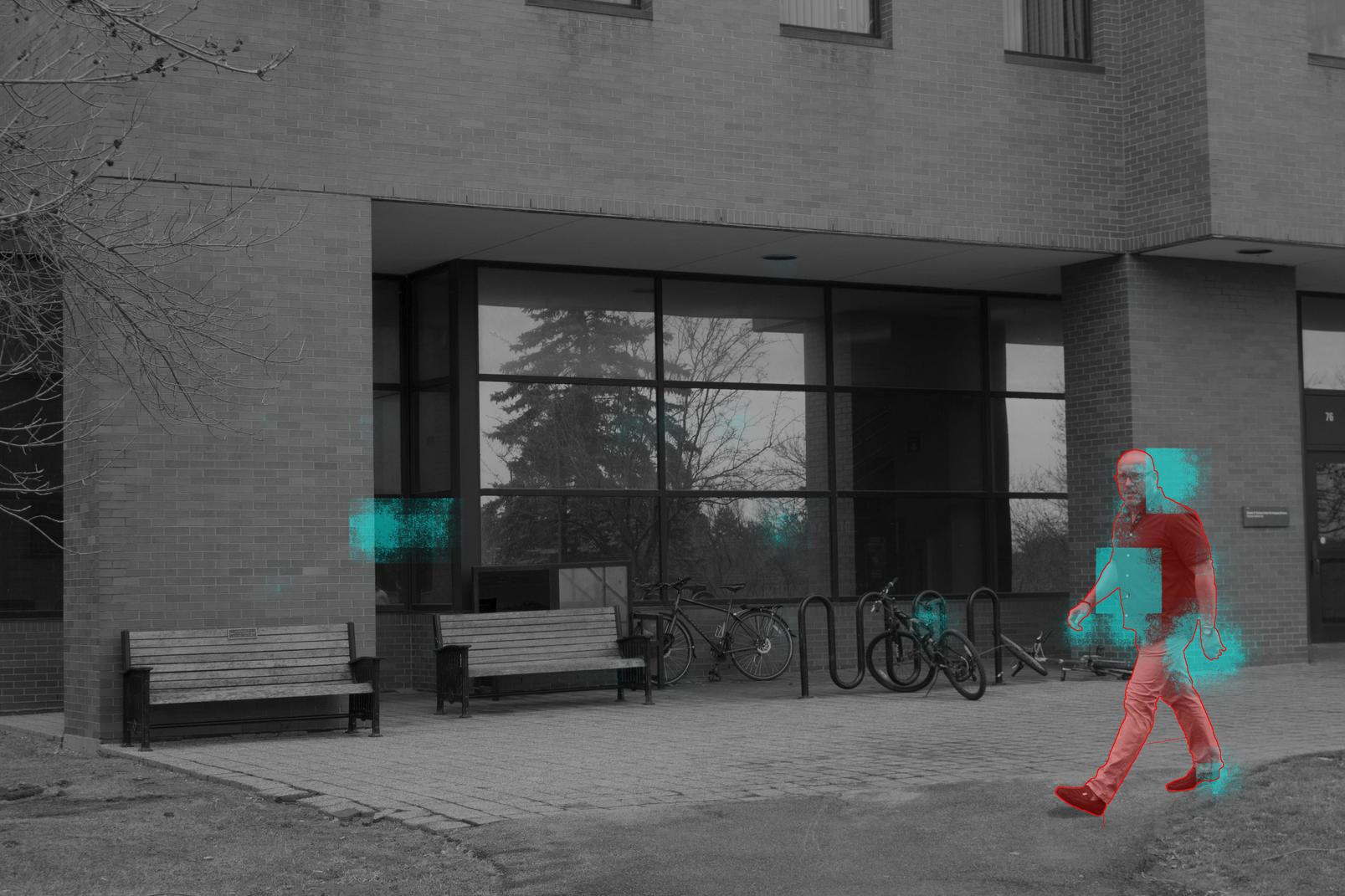}            &
    \includegraphics[width=0.240\linewidth]{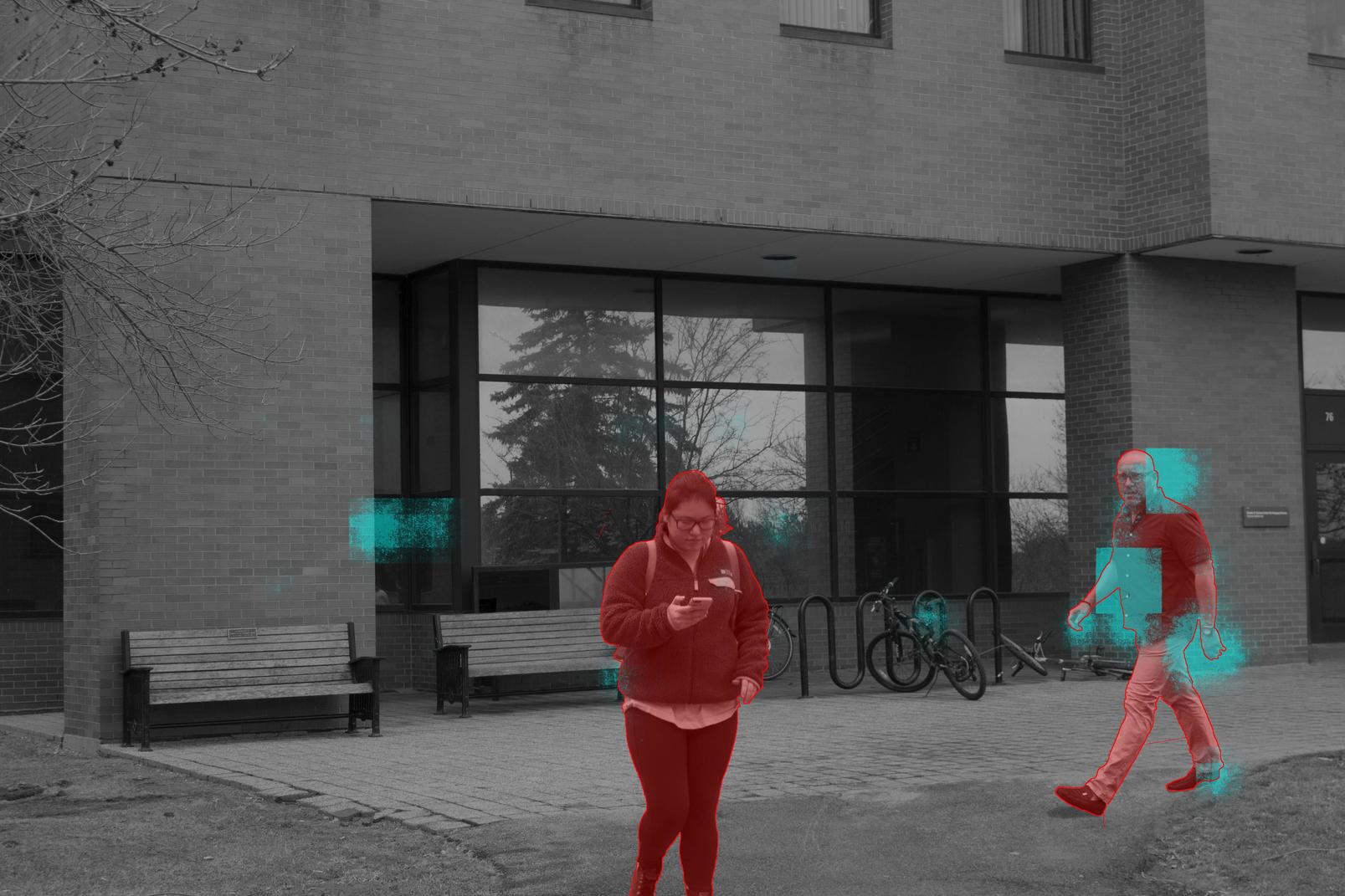}            &
    \includegraphics[width=0.240\linewidth]{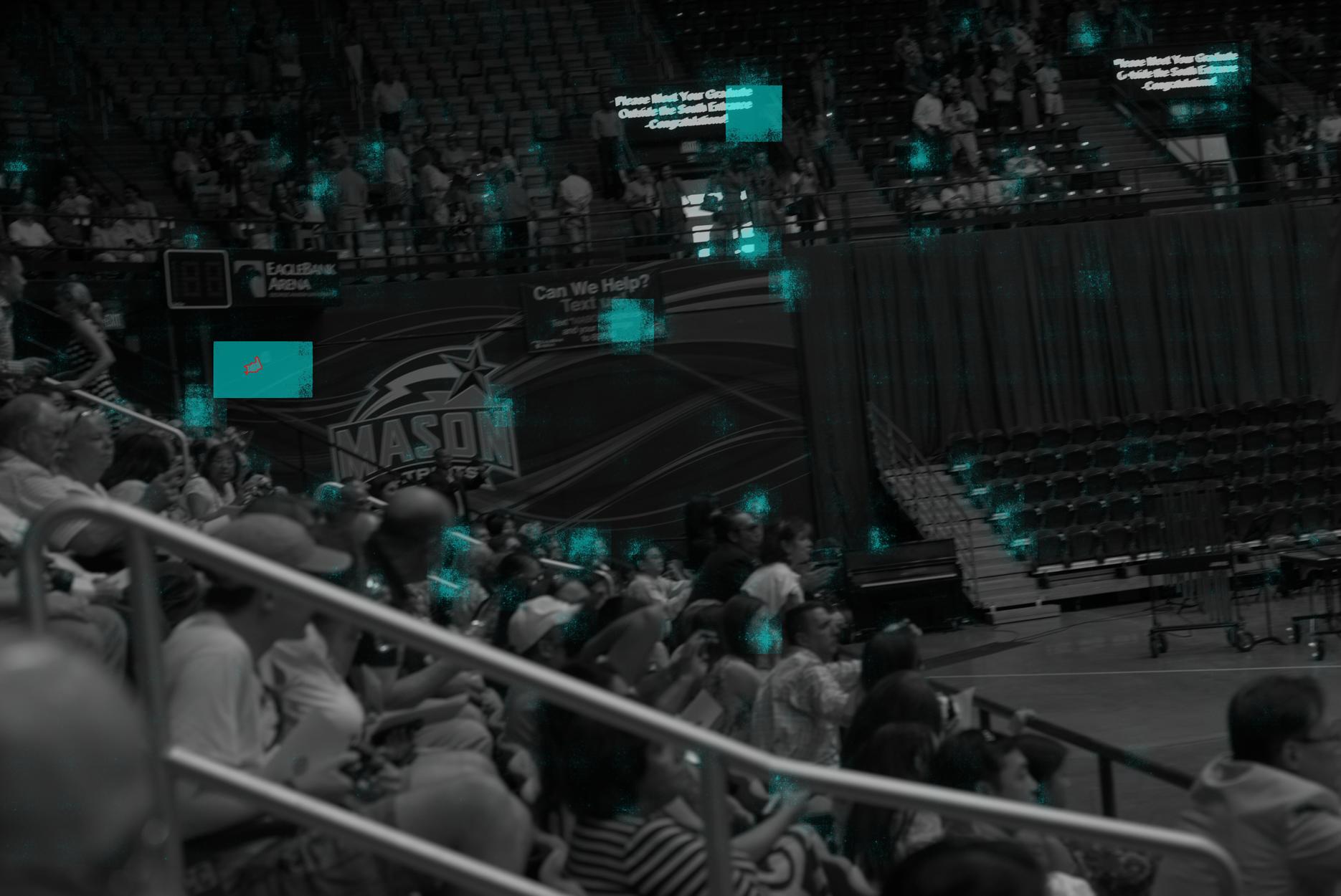}            &
    \includegraphics[width=0.214\linewidth]{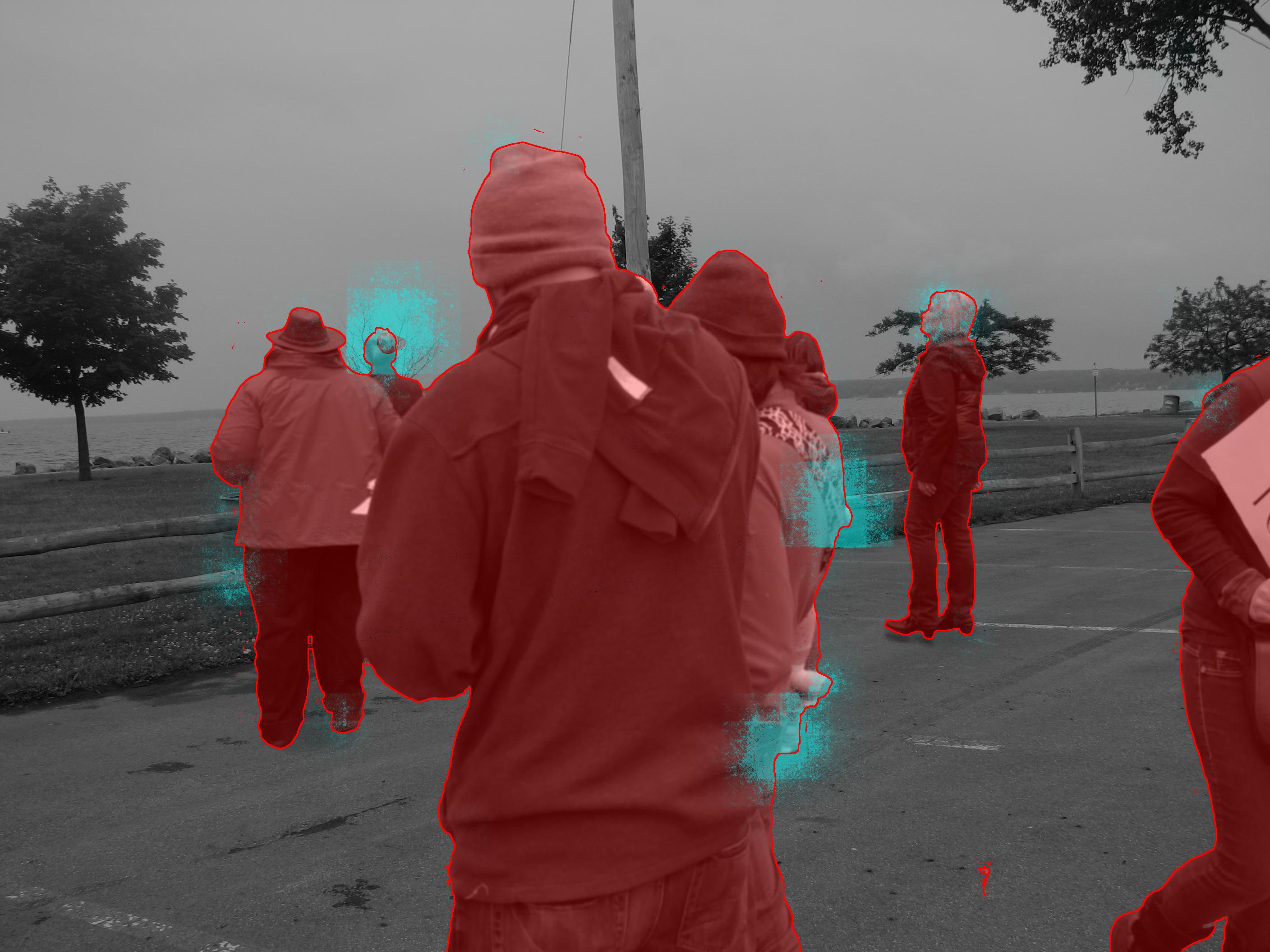}            \\
    \includegraphics[width=0.240\linewidth]{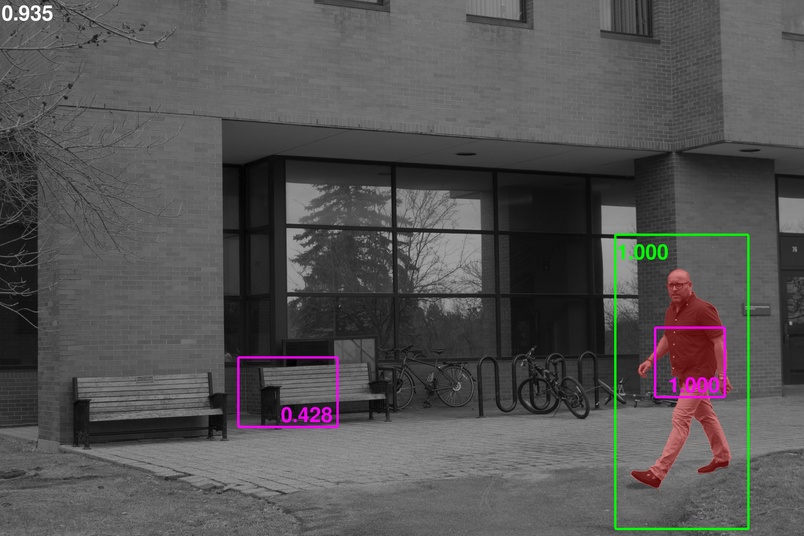} &
    \includegraphics[width=0.240\linewidth]{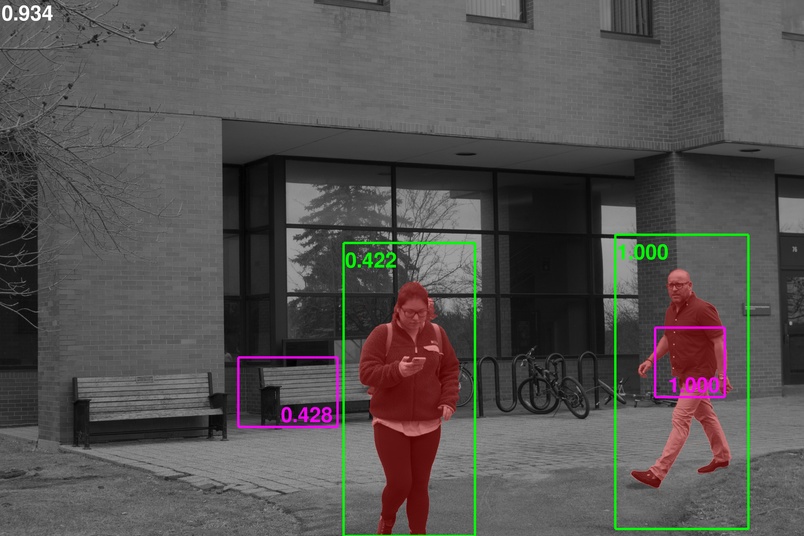} &
    \includegraphics[width=0.240\linewidth]{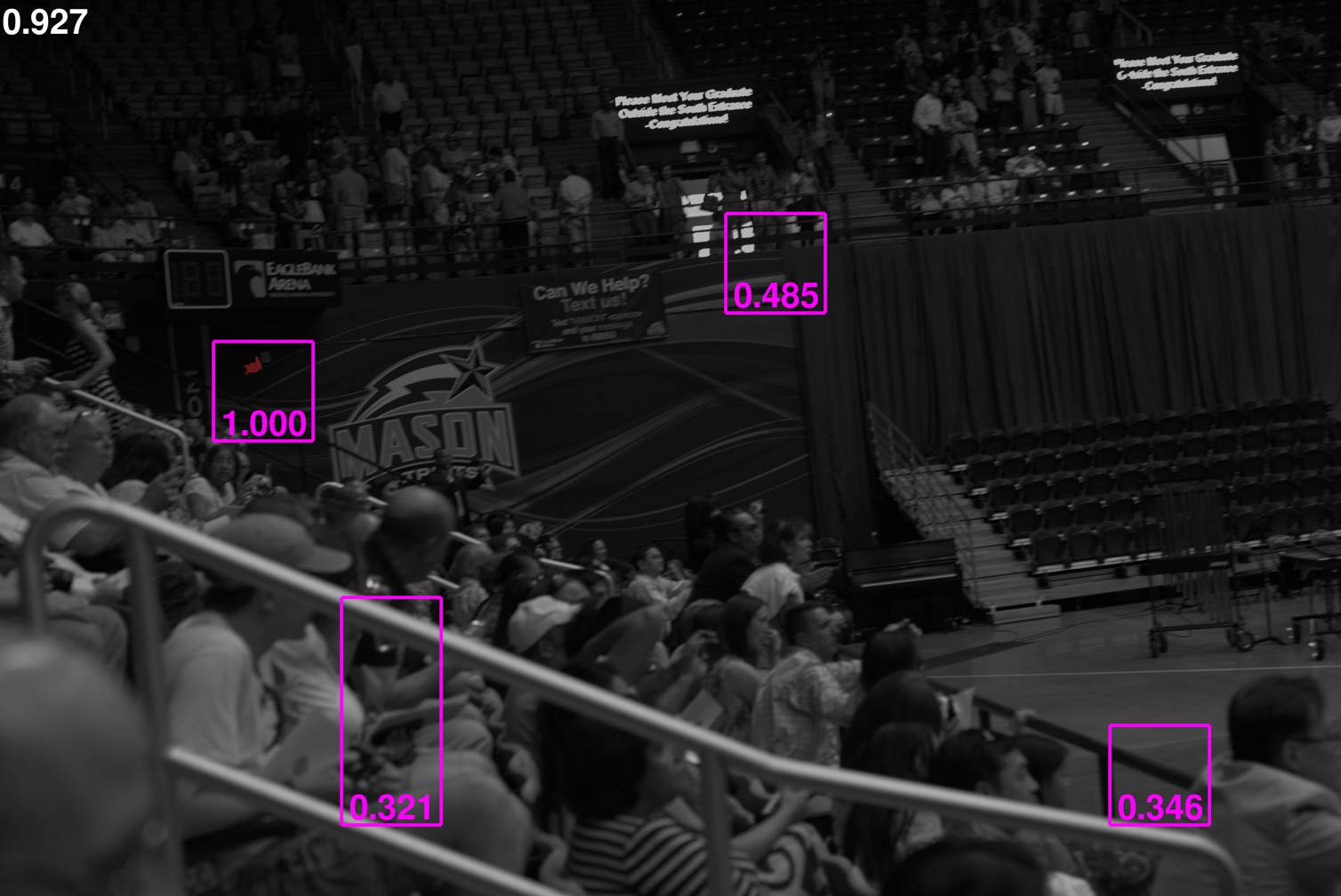} &
    \includegraphics[width=0.214\linewidth]{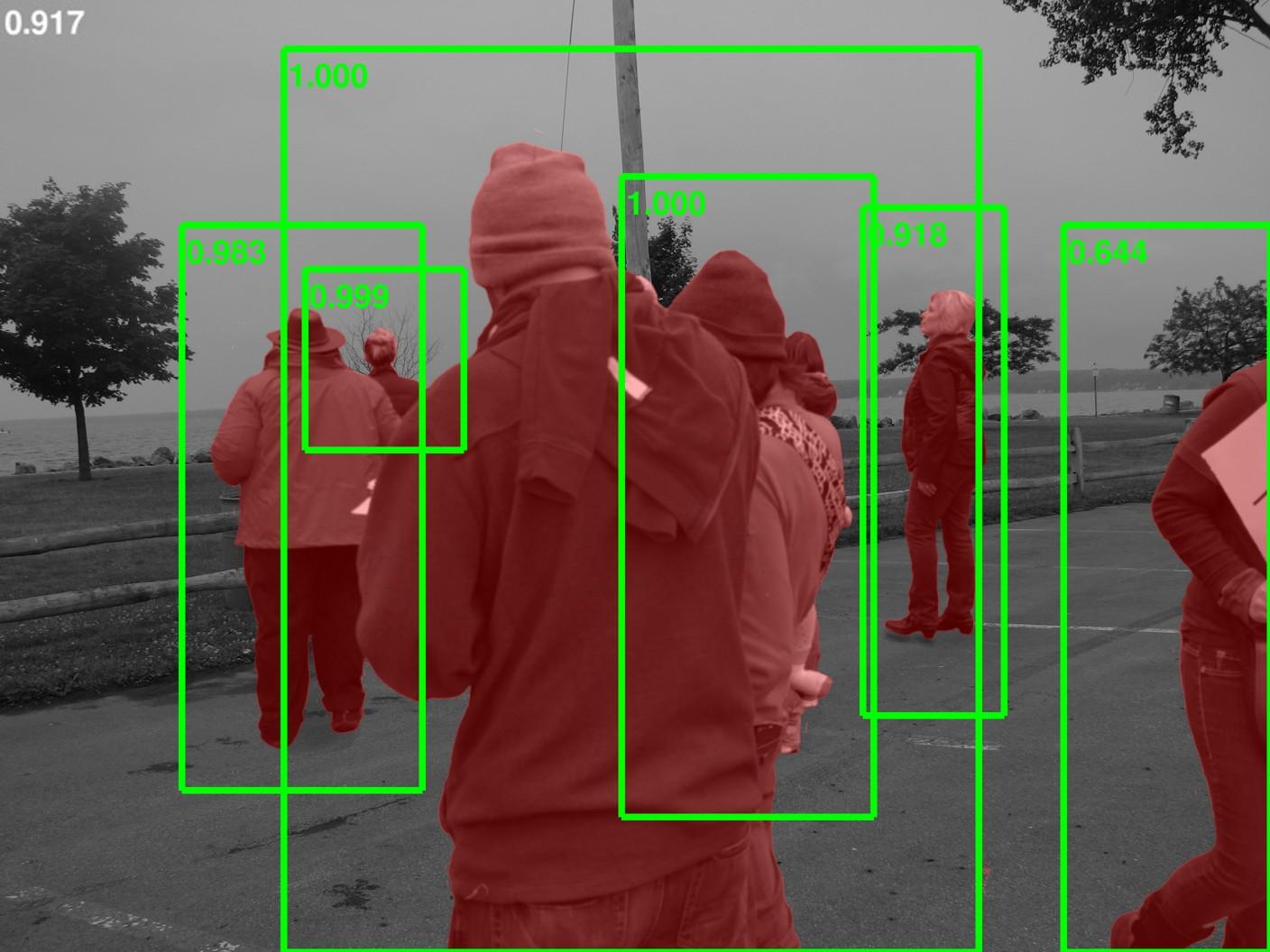} \\
    \end{tabular}
    \caption{Manipulated images from the NIST datasets (top) corresponding activation maps (middle) and ROI-based localization results (bottom) with hand-made (green) and automatic (magenta) box-shaped ROIs. Detection scores are shown on the top-left of each box.}
    \label{fig:MFC2018_boxes}
\end{figure*}

Fig.\ref{fig:MFC2018_boxes} shows some examples taken from the MFC2018 dataset.
Together with the original images (top) and activations maps (middle) it also shows (bottom) the scores obtained over the whole image (white number in the top-left corner) and on selected boxes (colored numbers).
The green boxes have been selected manually around possible subjects of interest,
while the magenta boxes are selected by our automatic procedure around the local maxima of the score.
In the first image, the man on the right has been spliced on the host background.
Here, the activation map provides strong hints on the possible manipulation, confirmed by a large image-level score (0.935).
However, an even larger score (1.000) is obtained when a ROI is correctly placed around the splicing.
The automatic procedure also selects a ROI roughly covering the splicing, with unitary score.
Another ROI is selected automatically in a pristine area in correspondence of a local maximum, but is has a rather low score (0.428).
In the second image, a further splicing has been added, the woman in the center.
Neither the activation map nor the automatic ROI selection procedure highlight this new subject.
So, we selected a ROI manually around this splicing, obtaining a rather low score.
Exploiting the side information provided with the NIST datasets,
we investigated on this splicing, to discover that the inserted object had been acquired with the same camera model as the host image.
This fact reduces the discriminating power of the noiseprint input, justifying in hindsight such result.
In the third image, the only manipulation is a tiny inpainted region.
Here, a supervised selection makes no sense, since the manipulated region does not correspond to any semantic object.
However, the manipulation is nicely localized through the automatic procedure, with unitary score, unlike other candidate ROIs characterized by low scores.
The last image shows an opposite case, with many large, semantically relevant, objects spliced on the host image.
To avoid cluttering the image, we now show only the supervised ROIs and the corresponding scores, which are very large in all cases.

\begin{figure}[t]
	\centering
	\setlength{\tabcolsep}{1mm}
	\begin{tabular}{cccc}
    \includegraphics[width=0.45\linewidth]{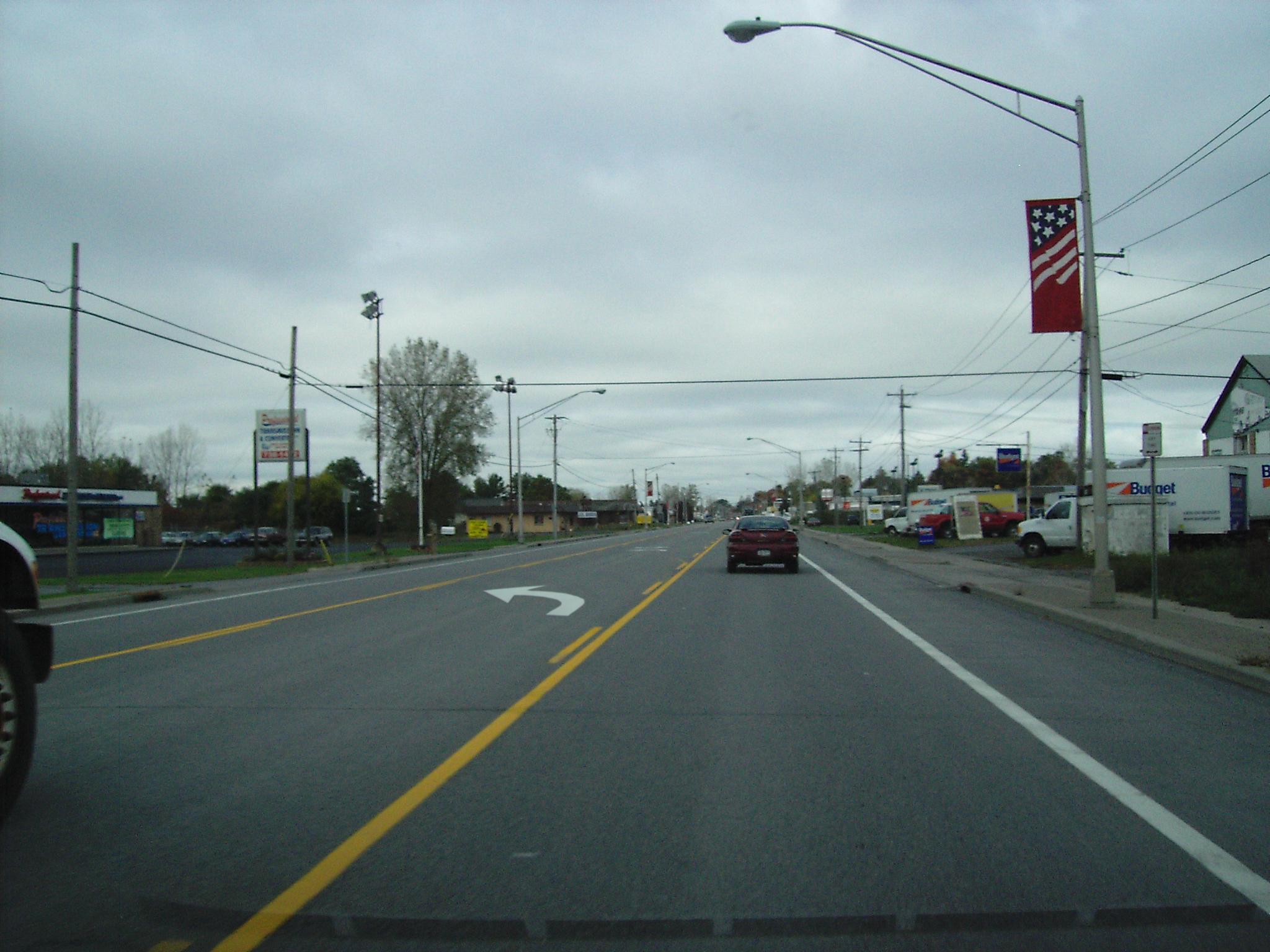}     &
    \includegraphics[width=0.45\linewidth]{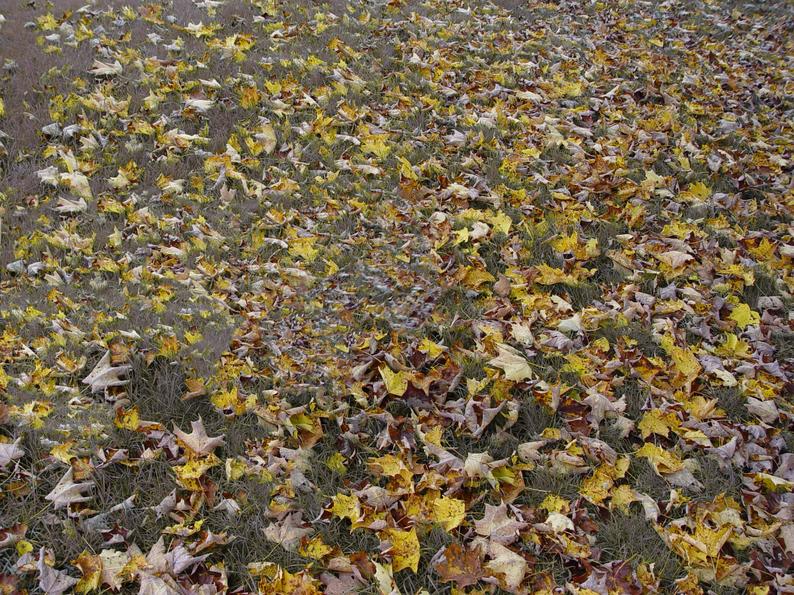}     \\
    \includegraphics[width=0.45\linewidth]{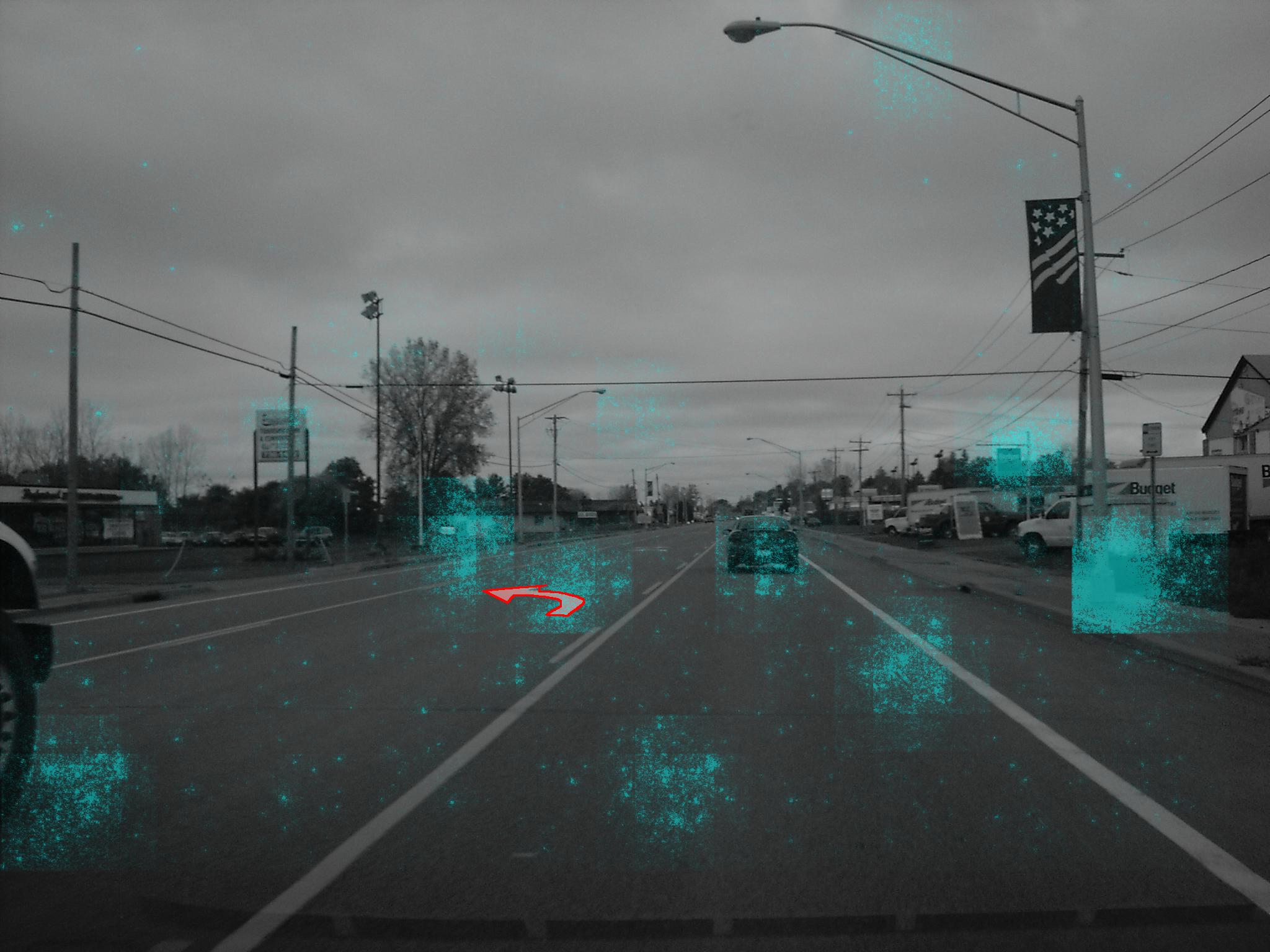}        &
    \includegraphics[width=0.45\linewidth]{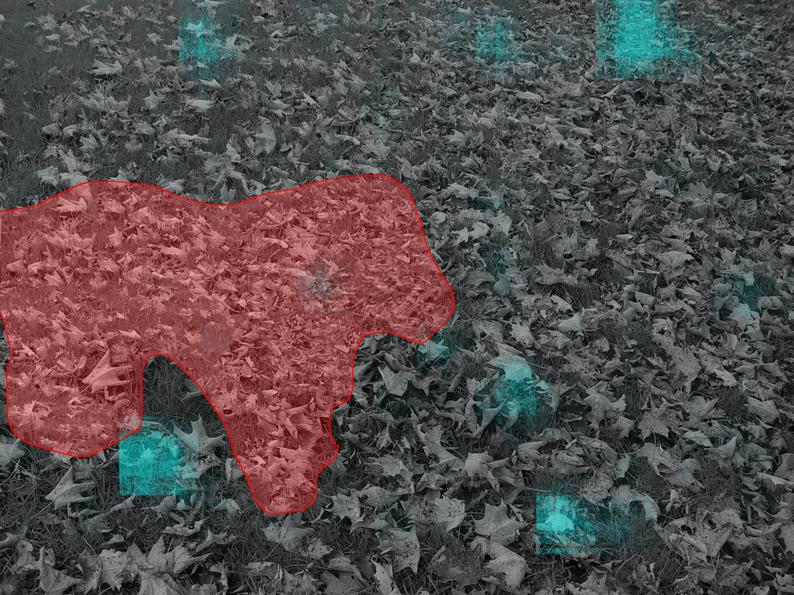}        \\
    \includegraphics[width=0.45\linewidth]{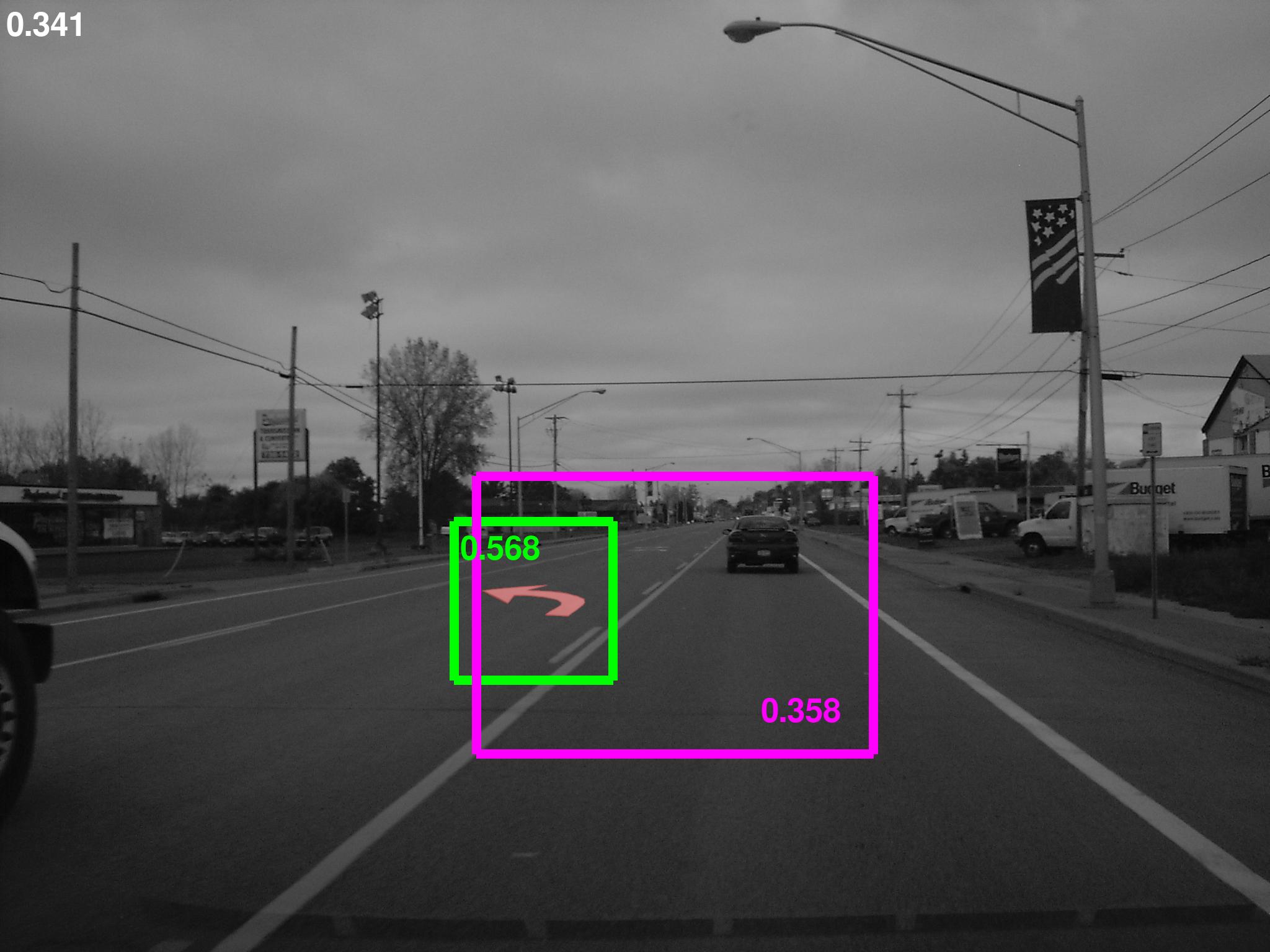} &
    \includegraphics[width=0.45\linewidth]{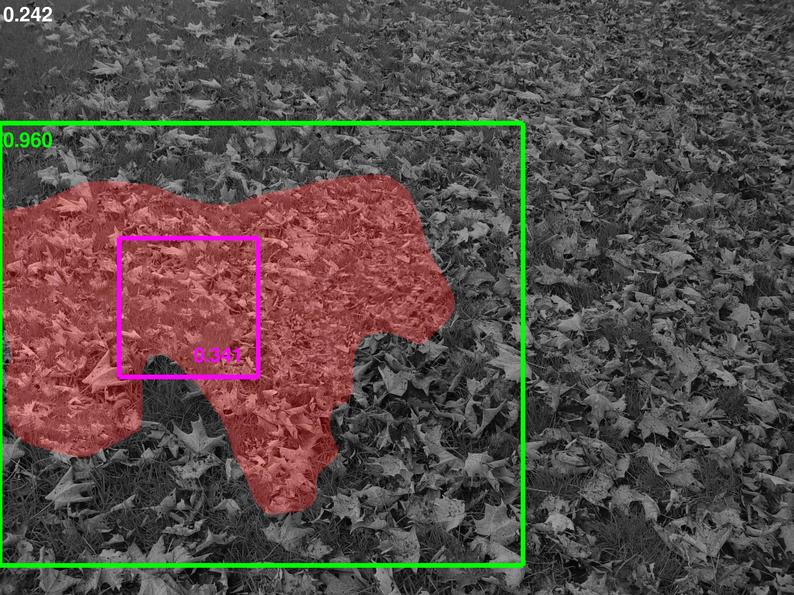} \\
    \end{tabular}
    \caption{Examples of missed detection from the NIST datasets.}
    \label{fig:MFC2018_boxes_bad}
\end{figure}

To complete this visual inspection of results, it is fair to show, in Fig.\ref{fig:MFC2018_boxes_bad}, some counter-examples where the proposed framework fails to detect the manipulation.
Reasons for failure are not always obvious.
In these cases, they may be related to the absence of texture in the spliced object (right) or the strongly textured host image (right) which may hide the discriminating information.
Note that in the image on the right, a well-placed ROI would allow detection, but there is no semantic hint to select it.

\section{Conclusion}
We proposed a new CNN-based framework for image forgery detection.
Thanks to suitable architectural solutions, it allows one to process jointly information gathered at full-resolution from the whole image.
Moreover, the framework can be trained end-to-end based only on weak (image-level) supervision.
We proved the effectiveness of this solution by extensive performance analysis on forensic datasets widespread in the community.
A large performance gain is observed in all cases with respect to all reference methods.
In addition, the framework can be also recast to provide localization information, both in supervised and unsupervised modality.

Despite the very promising results, there is still much room for improvement.
In particular, better forms of pooling should be considered to preserve long-range spatial relationships in the aggregation phase.
Moreover, image and object semantics should be taken into account to complement the low-level information analyzed by the current framework.
Work is already under way along these paths.

\section*{Acknowledgment}
This material is based on research sponsored by the Air Force Research Laboratory and the Defense Advanced Research Projects Agency under agreement number FA8750-16-2-0204.
The U.S.Government is authorized to reproduce and distribute reprints for Governmental purposes notwithstanding any copyright notation thereon.
The views and conclusions contained herein are those of the authors and should not be interpreted as necessarily representing the official policies or endorsements, either expressed or implied,
of the Air Force Research Laboratory and the Defense Advanced Research Projects Agency or the U.S. Government.

\balance
\bibliographystyle{IEEEtran}
\bibliography{refs}

\end{document}